\documentclass{article}

\usepackage[preprint]{neurips_2026}

\usepackage[utf8]{inputenc} % allow utf-8 input
\usepackage[T1]{fontenc}    % use 8-bit T1 fonts
\usepackage{hyperref}       % hyperlinks
\usepackage{url}            % simple URL typesetting
\usepackage{booktabs}       % professional-quality tables
\usepackage{amsfonts}       % blackboard math symbols
\usepackage{nicefrac}       % compact symbols for 1/2, etc.
\usepackage{microtype}      % microtypography
\usepackage{xcolor}         % colors

\usepackage{graphicx} 
\usepackage{subcaption}
\usepackage{amsmath}
\newcommand{\spm}{\scalebox{0.7}{$\pm$}}
\usepackage{multirow}
\usepackage{comment}

\title{Neural Operator Processes for Probabilistic Operator Learning under Partial Observations}

\author{%
  Jos\'e Miguel Lara Rangel \quad \quad Serge Guillas \\
  University College London \\
  London, United Kingdom \\
}

\begin{document}

\maketitle

\begin{abstract}
Neural operators learn mappings between function spaces, but are typically developed with dense input--output training fields and fully observed inputs at inference. Many scientific problems require instead predicting solution fields from sparse, irregular, or partial observations under uncertainty. We introduce Neural Operator Processes (NOPs), a framework that unifies neural-process conditioning with neural-operator decoding to predict full output fields from limited context. NOPs condition on sparse joint input--output observations and support deterministic and probabilistic prediction within a shared encoder--decoder architecture. We study two conditioning strategies, convolutional pooled summaries and query-aligned attention, and analyze how their interaction with latent stochastic variables depends on PDE geometry. Across function regression and three PDE benchmarks, we find that sparse conditional operator learning is viable and can match dense-grid behavior in several regimes, that preserving local context-query geometry is essential in non-periodic settings but less so in spectrally smooth periodic regimes, and that uncertainty-aware operator learning succeeds when latent conditioning complements rather than overwrites the local geometric pathway. These results provide a basis for probabilistic operator learning under partial observations and help bridge operator learning and probabilistic meta-learning in function space. 
\footnote{Code release is planned upon publication.}
\vspace{-0.4em}
\end{abstract}

\section{Introduction}
\label{sec:intro}

Learning mappings between function spaces has become a central problem in scientific machine learning.  In many applications, the goal is not to learn a finite-dimensional input--output map, but an operator that maps one field to another, such as a coefficient field to the solution of a partial differential equation~\citep{kovachki2024operator, li2020neuraloperator}. Neural operators (NOs), particularly architectures such as the Fourier Neural Operator (FNO), have shown a strong performance in this regime and established operator learning as an effective paradigm for surrogate modeling, simulation acceleration, and PDE-based inference~\citep{li2021fno, lu2021deeponet}. However, most NO methods are developed and evaluated in dense-grid settings that assume fully
paired input--output fields are available across the domain. This is more favorable than many realistic scenarios, such as data assimilation, where the target physical response is only known through sparse, irregular measurements, even if the underlying domain properties or forcing terms are known densely~\citep{cheng2023vivid, farchi2024nida}.

A second limitation is that uncertainty quantification remains underdeveloped in operator learning, especially under sparse observations. Many scientific settings require full-field prediction from only a small, irregular set of response measurements, where uncertainty reflects not just pointwise noise but task ambiguity induced by limited context. Existing approaches are often model-external, rely on ensembles or post-hoc calibration, or remain tied to dense input-output formulations~\citep{bulte2025pno, magnani2022abno, zhang2023metano}.

Neural Processes (NPs) offer a complementary perspective by learning distributions over functions from sparse context sets via amortized latent-variable inference, enabling rapid adaptation to new tasks without per-task optimization and providing predictive uncertainty~\citep{garnelo2018np,lara-rangel2025isanp,jha2022npfsurvey}. However, standard NPs are designed primarily for pointwise function regression and have neither been adapted nor widely applied to structured function-to-function mappings that demand dense PDE-scale decoding.

In this work, we introduce the \emph{Neural Operator Process} (NOP), a framework for learning distributions over operators under partial observations that combines the set-conditioned inference of NPs with the structured operator decoding of NOs, aiming to leverage the strengths of both paradigms. Rather than assuming access to densely paired input-output fields, NOPs condition on a sparse context set and predict the solution field over a dense query grid. This yields an episodic formulation of operator regression in which both the number and geometry of observations may vary across tasks. The framework naturally supports both deterministic and probabilistic variants, enabling a unified study of reconstruction accuracy, latent conditioning, and uncertainty quantification.

We study two complementary conditioning mechanisms: a global pooling pathway~\citep{gordon2020convcnp}, which compresses contextual information into a task-level representation, and a query-aligned attention pathway~\citep{kim2019anp}, which preserves local geometric structure. We extend both to probabilistic variants by introducing a global latent variable that captures task-level uncertainty while retaining the structured operator decoder. We examine how these pathways behave across PDE regimes and between deterministic and uncertainty-aware prediction.

Our experiments span four benchmark families with complementary roles. A Gaussian-process regression stage serves as a probabilistic diagnostic for sparse conditioning, latent inference, and uncertainty quantification. The PDE experiments then cover Burgers as a smooth periodic setting aligned with Fourier decoding, Darcy as a non-periodic and geometry-sensitive elliptic problem, and Navier--Stokes as a periodic but dynamically richer two-dimensional regime. Our results show that sparse conditional operator prediction is viable even with only a small fraction of joint observations, that the importance of preserving local context--query geometry is strongly regime-dependent and can become decisive in geometry-sensitive settings, that the framework provides meaningful uncertainty quantification, and that probabilistic operator learning is most reliable when latent conditioning complements rather than overrides the local geometric pathway. We complement these main results with controlled ablations on observation geometry, test-time context scaling, train-evaluation mismatch, cross-resolution transfer, local-conditioning design, and computational cost.

More broadly, our results suggest that operator learning and probabilistic meta-learning can be unified within a common framework that accounts jointly for structure, sparsity, uncertainty, adaptation, and probabilistic prediction in function space. Our main contributions are:

\begin{itemize}
    \item We formulate operator learning under partial observations as context-conditioned prediction in function space and propose Neural Operator Processes as a corresponding framework. 
    \item We introduce deterministic and probabilistic NOP architectures sharing an encoder--decoder pipeline, with different mechanisms for context aggregation and latent uncertainty.
    \item We identify an empirical distinction between global and local conditioning: geometry-preserving attention is especially valuable in geometry-sensitive settings, while the preferred pathway depends on the inference objective.
    \item We validate these findings through systematic experiments and ablations, showing that NOPs match or surpass dense-grid neural operators while observing only a small fraction of the response field (under 25\% in 1D and under 7\% in 2D), provide meaningful uncertainty estimates, and remain computationally competitive at inference.
\end{itemize}

\begin{figure}[t]
    \centering
    \scalebox{0.97}{%
    \begin{minipage}[c]{0.262\textwidth}
        \centering
        \includegraphics[width=\linewidth]{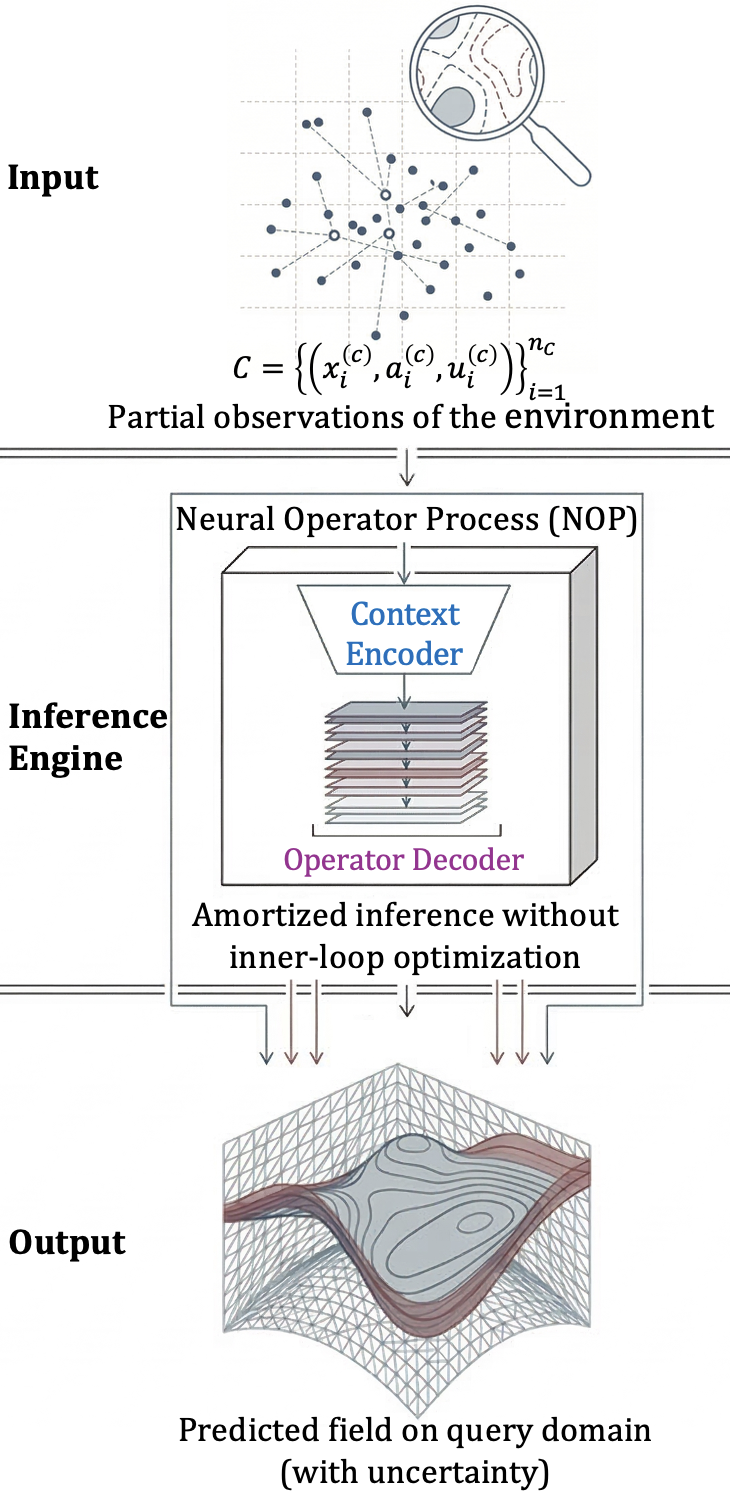}
        \subcaption{NOP framework.}
        \label{fig:nop_overview}
    \end{minipage}
    \hfill
    \begin{minipage}[c]{0.73\textwidth}
        \centering
        \includegraphics[width=\linewidth]{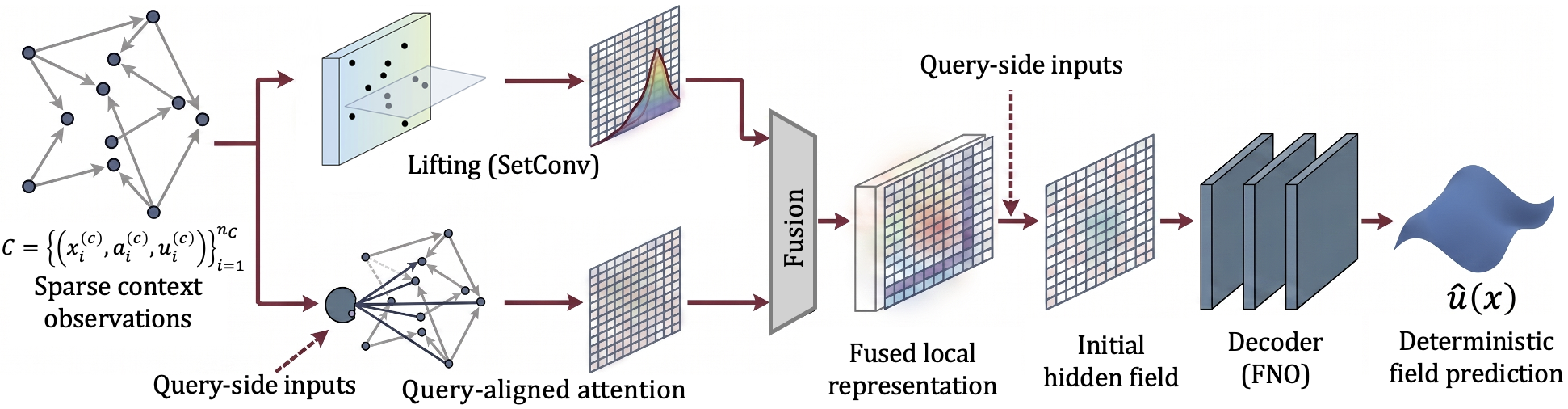}
        \subcaption{Deterministic NOP architecture.}
        \label{fig:nop_v0}
        \vspace{0.8em}
        \includegraphics[width=\linewidth]{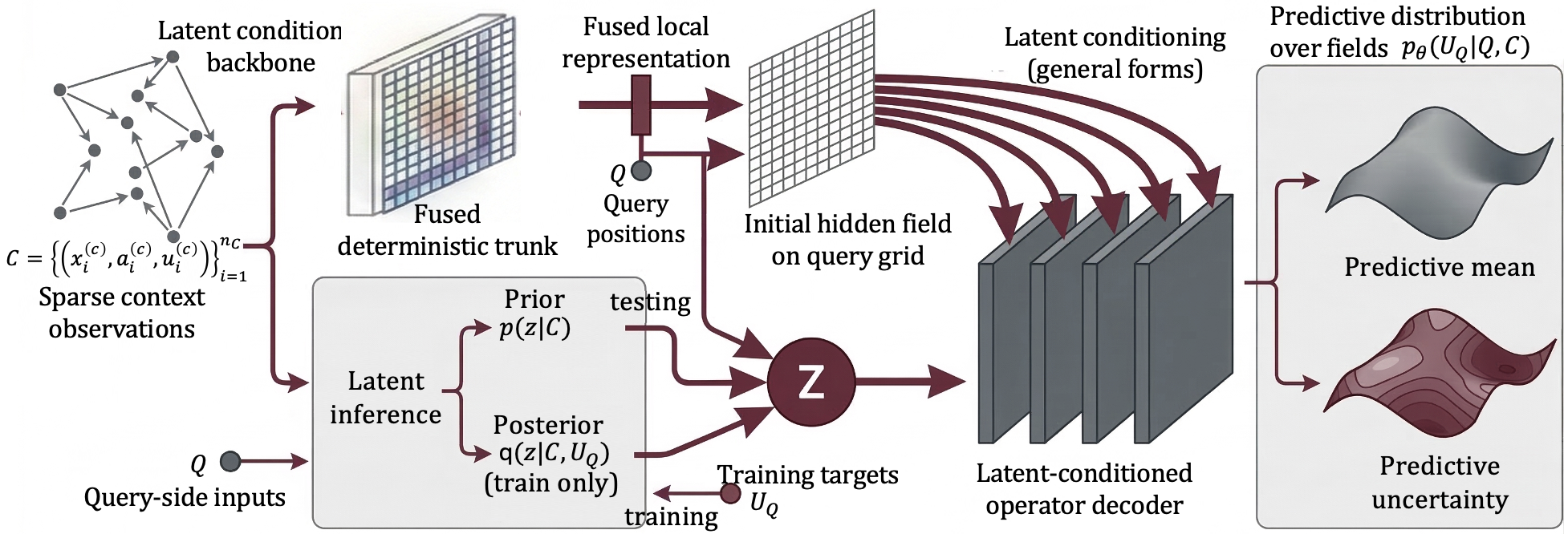}
        \subcaption{Probabilistic NOP architecture.}
        \label{fig:nop_prob}
    \end{minipage}
    }
    \caption{\textbf{NOP Pipeline.} (a) NOPs condition on a sparse context set of joint observations to infer the mapping from a given input field to its dense target solution. (b) The deterministic NOP constructs a local representation followed by an NO decoder. (c) The probabilistic NOP augments the conditional backbone with global latent inference, yielding a predictive distribution over output fields.}

    \label{fig:nop_main}
\vspace{-1em}
\end{figure}

\section{Background and Related Work}
\label{sec:background}

\subsection{Meta-Learning and Neural Processes}

Meta-learning studies how to use experience from related tasks to adapt quickly to a new task from limited data~\citep{finn2017maml}. Given a context dataset $C$, rather than fitting a single predictor over pooled data, meta-learning trains over related tasks to learn the map $C \mapsto f(\cdot; C)$, that is, an inference mechanism that adapts from a small context set~\citep{garnelo2018cnp}.

Neural Processes instantiate this view by learning a direct mapping from a sparse context set $C = \{(x_i, y_i)\}_{i=1}^{n_C}$ to a predictive distribution $p_\theta(Y_T \mid X_T, C)$ over function values at target inputs~\citep{ jha2022npfsurvey}. The context is treated as a \emph{set} with variable cardinality and no canonical ordering, requiring permutation-invariant processing~\citep{zaheer2017deepsets, garnelo2018cnp}.

In conditional NPs, context pairs are aggregated through a commutative operator to form a global summary, and a decoder conditioned on it outputs pointwise Gaussian predictions~\citep{garnelo2018cnp}. Latent NPs extend this framework with a global stochastic latent variable $z$, yielding predictions by integrating over its context-conditioned distribution. This allows the model to represent coherent task-level uncertainty shared across targets~\citep{garnelo2018np}. 

A known limitation is that aggressive global aggregation can discard local query-relevant structure ~\citep{jha2022npfsurvey}. Attention-based variants address it by allowing each target to attend directly to the context set ~\citep{kim2019anp}, while convolutional variants lift sparse contexts into continuous representations via kernelized set convolutions before convolutional processing ~\citep{gordon2020convcnp}.

\subsection{Neural Operators}

Neural Operators (NOs) aim to learn mappings between function spaces, rather than between finite-dimensional vectors~\citep{kovachki2024operator}. By targeting the underlying operator itself, instead of a single discretization, they can in principle generalize across resolutions~\citep{li2020neuraloperator}. Particularly, the FNO~\citep{li2021fno} parameterizes the operator through spectral convolution layers:
\[
v_{\ell+1}(x) = \sigma\!\left(W_\ell v_\ell(x) + \mathcal{F}^{-1}\!\big(R_\ell(k)\,\mathcal{F}(v_\ell)(k)\big)(x)\right),
\]
where $\mathcal{F}$ is the Fourier transform, $R_\ell(k)$ is a learned spectral multiplier truncated to finite modes, $W_\ell$ is a pointwise linear map, and $\sigma$ is a nonlinearity. This gives FNO a strong inductive bias toward globally coupled operators and is especially effective in periodic or spectrally smooth settings.

NOs are typically trained on fully paired dense input–output fields. Recent probabilistic and meta-learned variants~\citep{zhang2023metano, magnani2022abno, bulte2025pno} add uncertainty but still assume fully observed fields. We consider a different, complementary setting where the number and placement of context observations are part of the conditioning problem, so NPs serve as a probabilistic set-encoder while NOs provide scalable, structured decoders for function-to-function mappings. The main challenge is to unify them to infer a complete output field from a small context set, yield coherent uncertainty quantification, and preserve inductive biases of NOs for dense structured decoding.

\section{Neural Operator Processes}
\label{sec:nop}

\subsection{Problem Setup}

Let $\Omega \subset \mathbb{R}^{d_x}$ be the spatial domain and $\mathcal{G}: a \mapsto u$ an operator mapping an input field $a$ to an output field $u$. Each task corresponds to an underlying input–output field pair  $(a(\cdot), u(\cdot))$, for example a coefficient field and its PDE solution, or an initial condition and its evolved state. 

For a given task, we sample a context set $C = \{(x_i^{(c)}, a_i^{(c)}, u_i^{(c)})\}_{i=1}^{n_C}$, \(x_i^{(c)} \in \Omega\), and define a query set $Q = \{(x_j^{(q)}, a_j^{(q)})\}_{j=1}^{n_Q}$ with target outputs $U_Q = \{u_j^{(q)}\}$. Although each task is generated from a discretized input--output pair, during training and evaluation the model is exposed only to sparse pointwise observations. The query set typically corresponds to a structured grid compatible with the NO decoder. The query-side input $a_Q$ is provided on the grid; the missing information concerns the sparse availability of joint $(a, u)$ observations.

This induces a conditional operator-inference problem from sparse observations. Given a sparse context set and query-side information, the model must infer the task-specific functional relation and predict the output field over the query domain. In the deterministic setting, this corresponds to learning $\hat{U}_Q = f_\theta(Q, C)$; in the probabilistic setting, the goal is to model $p_\theta(U_Q \mid Q, C)$.

\subsection{Model Overview}

As shown in Figure~\ref{fig:nop_main}, NOP proceeds by encoding the sparse context set into a context-conditioned representation \(R_C(x)\) evaluated at query locations, combining it with query-side input information to form an initial hidden field \(v_0(x)\), and then decoding it with a NO decoder \(\mathcal{D}_\theta\) to obtain the final prediction \(\hat{u}(x)\). This general pipeline can be instantiated in a deterministic or probabilistic version.

\subsubsection{Deterministic Neural Operator Process}

The deterministic NOP defines the core conditional architecture. A key design choice is how sparse context information is represented before decoding. To bridge the set-based view of NPs with the continuous-domain view of NOs, we employ a set-to-function lifting step~\citep{gordon2020convcnp}. We consider two complementary pathways, which may be used separately or fused before decoding.

\paragraph{SetConv Conditioning.} We construct a continuous representation field $R_C(x) = \mathrm{SetConv}_\phi(C; x)$, evaluable at arbitrary query locations $x \in \Omega$. Let $w_\phi$ be a kernel and $\epsilon > 0$ a stabilizer:
\begin{equation*}
R_C(x) = \Bigg[\sum_{i=1}^{n_C} w_\phi(x - x_i^{(c)}), \quad
\frac{\sum_{i=1}^{n_C} [a_i^{(c)}, u_i^{(c)}]\, w_\phi(x - x_i^{(c)})}
{\sum_{i=1}^{n_C} w_\phi(x - x_i^{(c)}) + \epsilon}\Bigg].
\end{equation*}
This lifts an irregular sparse context set into a continuous field compatible with the query domain for the operator decoder, while separately encoding where context support is concentrated (density) and what local field values are observed (signal). We implement $w_\phi$ as an RBF kernel. In practice, additional positional or geometric features can be included to better account for domain geometry.

\paragraph{Query-Aligned Attention Conditioning.} To capture context--query interactions beyond kernel proximity, we introduce a query-aligned attention pathway that models which context points are most relevant to each query~\citep{kim2019anp, lee2019settransformer}. Let $h_i^{(c)} = \phi_{\mathrm{ctx}}(x_i^{(c)}, a_i^{(c)}, u_i^{(c)})$ denote a learned context embedding and $q(x) = \phi_{\mathrm{qry}}(x, a(x))$ a query embedding:
\begin{equation*}
r_{\mathrm{att}}(x) = \sum_{i=1}^{n_C} \alpha_i(x)\, h_i^{(c)}, \qquad \alpha_i(x) = \frac{\exp(\omega_\theta(h_i^{(c)}, q(x)))}{\sum_{j=1}^{n_C}\exp(\omega_\theta(h_j^{(c)}, q(x)))},
\end{equation*}
where $\omega_\theta$ is a learned compatibility function. This yields a target-specific context summary that preserves local geometric structure.

\paragraph{Fused Local Representation.} The two pathways can be combined into a fused representation $r_{\mathrm{fuse}}(x) = \phi_{\mathrm{fuse}}(R_C(x), r_{\mathrm{att}}(x))$, implemented in this work as a pointwise MLP on concatenated inputs. This is combined with query-side information to form the initial hidden field $v_0(x) = P_\theta\big([x,\; a(x),\; r_{\mathrm{fuse}}(x),\; b(x)]\big)$, where $P_\theta$ is a pointwise projection and $b(x)$ denotes optional positional features (e.g., boundary-distance encodings). Both conditioning pathways cost $O(n_C n_Q)$ for context--query interactions, while the subsequent $L$-layer FNO decoder scales as $O(L n_Q \log n_Q)$, up to memory and channel constants.

\paragraph{Operator Decoder.} The resulting hidden field is passed through an NO decoder; we use an FNO-style decoder. After $L$ operator layers, a pointwise output head maps the final representation to the prediction, yielding $\hat{u}(x) = \phi_{\mathrm{out}}(v_L(x))$. To encourage accurate prediction of the full query field while remaining robust to scale differences across tasks, we train using a relative $L^2$ loss.

\subsection{Probabilistic Neural Operator Process}
 
We extend the deterministic architecture by introducing a latent-variable pathway to obtain $p_\theta(U_Q \mid Q, C)$ rather than a point estimate. A key design principle is that the probabilistic extension should complement, not replace, the local geometric conditioning.

\paragraph{Latent Prior and Posterior.} Let $z \in \mathbb{R}^{d_z}$ be a global latent variable capturing task-level uncertainty. We infer $z$ from a global summary of the context-induced representation, obtained by permutation-invariant aggregation over the structured query grid $Q_{\mathrm{ref}}$, $r_C = \mathrm{Agg}(\{R_C(x)\}_{x \in Q_{\mathrm{ref}}})$. The context-conditioned prior is $p_\theta(z \mid C) = \mathcal{N}(z; \mu_p(r_C), \mathrm{diag}(\sigma_p^2(r_C)))$. During training, an approximate posterior $q_\phi(z \mid C, U_Q)$ has access to target outputs through a representation $r_{C \cup Q^{\mathrm{obs}}}$ built from the union of the context and the training query points with known target values, using the same set-to-function construction as the prior.

\paragraph{Latent Conditioning and Predictive Distribution.} The latent $z$ modulates the deterministic backbone via broadcast concatenation with $v_0(x)$. A probabilistic head outputs $\mu_u(x)$ and $\sigma_u^2(x)$, and the predictive distribution is obtained by marginalizing the conditionally factorized likelihood:
\begingroup
\setlength{\abovedisplayskip}{0.4pt}
\setlength{\belowdisplayskip}{0.4pt}
\begin{equation*}
p_\theta(U_Q \mid Q,C)
=
\int
\prod_{j=1}^{n_Q}
\mathcal{N}\!\left(u_j^{(q)};\mu_u(x_j^{(q)}),\sigma_u^2(x_j^{(q)})\right)
p_\theta(z\mid C)\,dz .
\end{equation*} 
\endgroup
The resulting uncertainty has two components: a global task-level component from $z$, and a local predictive component from the output likelihood. In all probabilistic models we use input-stage latent injection, where $z$ enters only through $v_0$, as this yielded substantially more stable prior-conditioned inference than decoder-wide latent modulation in our latent-injection ablation (Appendix~\ref{app:latent_injection_ablation}).

%A probabilistic output head predicts $p_\theta(u(x)\mid z,Q,C) = \mathcal{N}(\mu_u(x),\sigma_u^2(x))$, and the full predictive distribution is obtained by marginalization:
%\begin{equation*}
%p_\theta(U_Q \mid Q, C) = \int p_\theta(U_Q \mid z, Q, C)\, p_\theta(z \mid C)\, dz.
%\end{equation*}

\paragraph{Training and Inference.} Training uses ELBO with linear \(\beta\)-scheduling, while test-time inference uses the prior \(z \sim p_\theta(z \mid C)\), approximated by Monte Carlo sampling:
\[
\mathcal{L}_{\mathrm{ELBO}} = \mathbb{E}_{q_\phi(z \mid C, U_Q)}[\log p_\theta(U_Q \mid z, Q, C)] - \beta\, \mathrm{KL}(q_\phi(z \mid C, U_Q) \| p_\theta(z \mid C)).
\]

\section{Experimental Setup}
\label{sec:setup}

\subsection{Benchmarks and Task Families}

We consider four benchmark families in a controlled progression to assess whether the probabilistic setup behaves coherently, whether sparse context-conditioned prediction works for PDEs, which conditioning is most effective, whether probabilistic extensions provide meaningful uncertainty without substantially degrading accuracy, and how behavior varies across regimes and geometries.

\paragraph{GP 1D regression.} Tasks sampled from GPs with RBF, Mat\'ern-5/2, and mixed kernels. This serves to verify that sparse conditioning, latent inference, and uncertainty quantification behave coherently. GP tasks are exposed through the same NOP interface used for PDEs.

\paragraph{Burgers 1D equation.} Viscous Burgers with periodic boundary conditions, mapping initial condition $u_0(x) \mapsto u_T(x)$. Smooth, periodic, and spectrally aligned with FNO.

\paragraph{Darcy 2D flow.} Elliptic operator on a non-periodic bounded domain, mapping permeability $a(x)$ to pressure $u(x)$. Boundary-sensitive and geometry-demanding to test local geometry preservation.

\paragraph{Navier--Stokes 2D.} Incompressible Navier--Stokes in vorticity formulation on a periodic domain ($\nu = 10^{-3}$, $T = 5.0$), mapping $\omega_0(x) \mapsto \omega_T(x)$. A canonical 2D dynamic benchmark that is periodic but spatially complex, occupying an intermediate position between Burgers and Darcy.

Each task is one episode, with full-query supervision and disjoint context and target sets, so at evaluation NOPs adapt to held-out response fields through sparse context without test-time optimization. We use \(1000/200/200\) tasks split across all datasets. Context sizes range from \(16\!-\!64\) points on Burgers (\(6.25\%\!-\!25\%\) of the 256-point grid), \(32\!-\!256\) on Darcy (\(\approx 0.78\%\!-\!6.25\%\) of \(64^2=4096\) points), and \(64\!-\!256\) on Navier-Stokes (\(\approx 1.56\%\!-\!6.25\%\) of 4096 points). Appendix ablations study additional context-sampling strategies, train--evaluation combinations, and cross-resolution settings.

\subsection{Models}

We evaluate four NOP variants that explore different aspects of the framework and learning setup. \textbf{DNOP} is the deterministic baseline with SetConv conditioning, and \textbf{PNOP} its probabilistic extension with input-stage latent injection. \textbf{DANOP} is the deterministic model with fused SetConv + attention conditioning, and \textbf{PANOP} its probabilistic extension. We also compare heteroscedastic (HE) and homoscedastic (HO) output heads. To provide a meaningful dense-input reference, we include an FNO adapted to the same tasks and output targets considered for NOPs, but operating on full fields. The appendix reports additional comparisons with other architectural variants and NO baselines.
 %We evaluate with rel-$L^2$ error, and include MC-NLL, empirical coverage, and average predictive interval width for probabilistic variants. These should be interpreted jointly as high coverage alone is not sufficient when achieved through excessively wide intervals, and low pointwise error does not by itself imply reliable uncertainty. All results are reported over 4 seeds unless noted.

\section{Results}
\label{sec:results}

\begin{figure*}[t]
\centering
\begin{minipage}{0.5\linewidth}
\centering
\includegraphics[width=\linewidth]{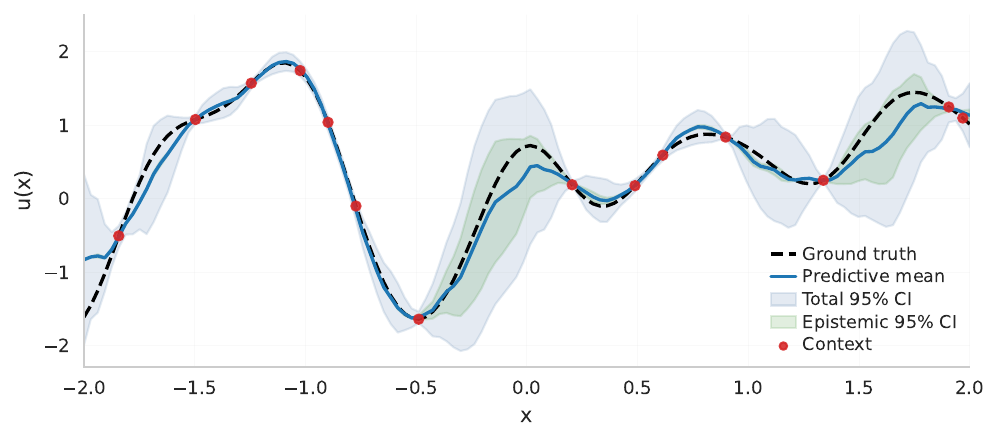}
\end{minipage}\hfill
\begin{minipage}{0.5\linewidth}
\centering
\includegraphics[width=\linewidth]{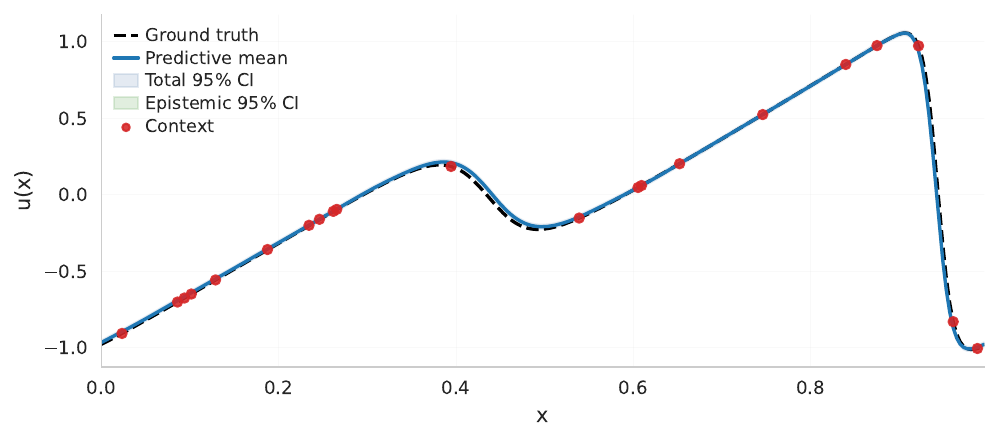}
\end{minipage}
\vspace{-0.5em}
\caption{\textbf{PNOP-HE behavior in 1D GP regression and Burgers.} On GP-RBF (left), the model tracks the function and predictive uncertainty increases in weakly constrained regions. Epistemic intervals are computed from the variance of predictive means across latent samples $z \sim p_\theta(z\mid C)$, while total intervals additionally include the predicted conditional likelihood variance. On Burgers (right), the main visible behavior is the accurate predictive mean; predictive intervals are small at the scale of the solution, with the variance decomposition shown in Figure~\ref{fig:burgers_he_ho_decomp}.}
\label{fig:prob_1d_gp_burgers}
\vspace{-1em}
\end{figure*}

\paragraph{NOP Behavior on GP regression.} Table~\ref{tab:gp_combined_main} shows that both architectures learn the underlying function. DANOP consistently improves on DNOP across all GP families, with larger gains on Mixed and Mat\'ern than on RBF. This suggests that preserving local context--query geometry is already useful in sparse function regression, even before considering PDEs. For the probabilistic variants, prior-mean performance remains reasonably close to posterior-mean performance, while zero-latent evaluation degrades sharply across all datasets. The latent pathway is active and not collapsed, and the prior--posterior gap remains modest. The uncertainty quantification behaves coherently and, as depicted in Figure~\ref{fig:prob_1d_gp_burgers}, predictions appropriately widen uncertainty in weakly constrained regions, consistent with coherent probabilistic behavior under sparse conditioning.

\begin{table}[t]
\centering
\setlength{\tabcolsep}{2pt}
\caption{Deterministic and probabilistic heteroscedastic variants rel-$L^2$ on GP benchmark with different latent inference modes. Mean values over 4 seeds with standard deviation. Lower is better.}
\label{tab:gp_combined_main}
\resizebox{\linewidth}{!}{%
\begin{tabular}{lcccccccc}
\toprule
& \multicolumn{2}{c}{\textbf{Deterministic}} & \multicolumn{2}{c}{\textbf{Prior-mean}} & \multicolumn{2}{c}{\textbf{Posterior-mean}} & \multicolumn{2}{c}{\textbf{Zero-latent}} \\
\cmidrule(lr){2-3} \cmidrule(lr){4-5} \cmidrule(lr){6-7} \cmidrule(lr){8-9}
\textbf{GP} & \textbf{DNOP} & \textbf{DANOP} & \textbf{PNOP-HE} & \textbf{PANOP-HE} & \textbf{PNOP-HE} & \textbf{PANOP-HE} & \textbf{PNOP-HE} & \textbf{PANOP-HE} \\
\midrule
RBF 
& $0.248 \spm 0.002$ & $0.243 \spm 0.002$ 
& $0.264 \spm 0.001$ & $0.265 \spm 0.000$ 
& $0.197 \spm 0.001$ & $0.197 \spm 0.002$ 
& $0.543 \spm 0.012$ & $0.568 \spm 0.013$ \\
Mixed 
& $0.352 \spm 0.001$ & $0.340 \spm 0.001$ 
& $0.359 \spm 0.000$ & $0.363 \spm 0.001$ 
& $0.295 \spm 0.001$ & $0.295 \spm 0.006$ 
& $0.648 \spm 0.001$ & $0.648 \spm 0.008$ \\
Mat\'ern 
& $0.411 \spm 0.001$ & $0.395 \spm 0.001$ 
& $0.419 \spm 0.001$ & $0.420 \spm 0.001$ 
& $0.352 \spm 0.001$ & $0.351 \spm 0.005$ 
& $0.710 \spm 0.008$ & $0.711 \spm 0.020$ \\
\bottomrule
\end{tabular}
}
\vspace{-1em}
\end{table}

\begin{table*}[t] 
\caption{Deterministic models mean rel-$L^2$ error over 4 seeds across PDEs, comparing standard uniform context, clustered context shift, and zero-shot (Coarse $\rightarrow$ Fine) transfer.}
\label{tab:det_main}
\centering
\setlength{\tabcolsep}{2pt}
\resizebox{1.00\linewidth}{!}{%
\begin{tabular}{l c cc cc cc}
\toprule
& \textbf{Dense Grid} & \multicolumn{2}{c}{\textbf{Uniform Context}} & \multicolumn{2}{c}{\textbf{Clustered Context}} & \multicolumn{2}{c}{\textbf{Zero-Shot}} \\
\cmidrule(lr){2-2} \cmidrule(lr){3-4} \cmidrule(lr){5-6} \cmidrule(lr){7-8}
\textbf{PDE} & \textbf{FNO} & \textbf{DNOP} & \textbf{DANOP} & \textbf{DNOP} & \textbf{DANOP} & \textbf{DNOP} & \textbf{DANOP} \\
\midrule
Burgers & $0.0050 \spm 0.0001$ & $0.0058 \spm 0.0001$ & $0.0047 \spm 0.0001$ & $0.0115 \spm 0.0003$ & $0.0059 \spm 0.0003$ & $0.0057 \spm 0.0001$ & $0.0048 \spm 0.0001$ \\
Darcy   & $0.0254 \spm 0.0003$ & $0.0242 \spm 0.0002$ & $0.0226 \spm 0.0008$ & $0.0176 \spm 0.0013$ & $0.0232 \spm 0.0008$ & $0.3316 \spm 0.0131$ & $0.3196 \spm 0.0167$ \\
NS      & $0.0343 \spm 0.0004$ & $0.0323 \spm 0.0030$ & $0.0326 \spm 0.0016$ & $0.0354 \spm 0.0002$ & $0.0336 \spm 0.0007$ & $0.2195 \spm 0.0033$ & $0.2031 \spm 0.0045$ \\
\bottomrule
\end{tabular}%
}
\vspace{-1em}
\end{table*}

\begin{figure*}[t]
\centering
\begin{minipage}{0.48\linewidth}
\centering
\includegraphics[width=\linewidth]{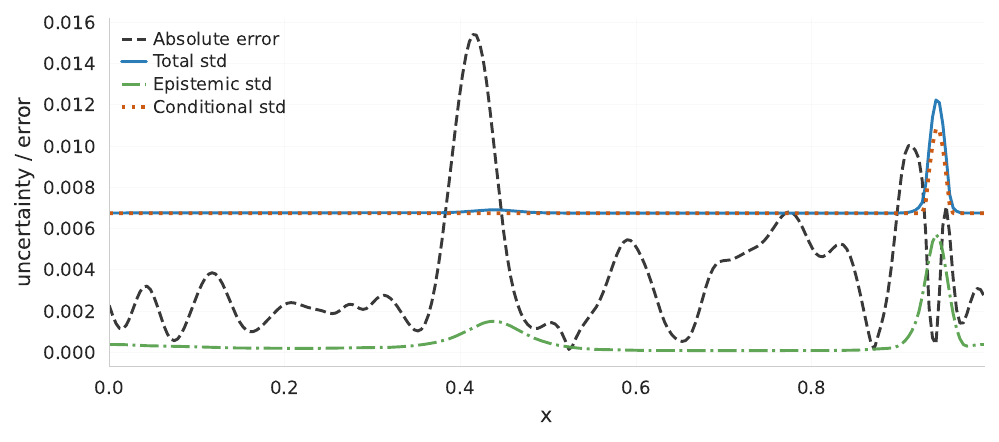}
\end{minipage}\hfill
\begin{minipage}{0.48\linewidth}
\centering
\includegraphics[width=\linewidth]{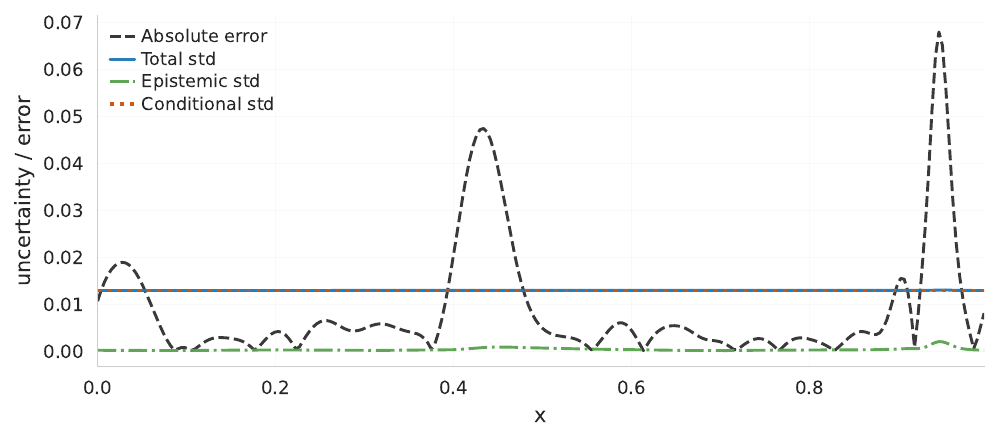}
\end{minipage}
\vspace{-0.5em}
\caption{\textbf{PANOP variance decomposition in Burgers.} The HE head (left) yields more spatially adaptive uncertainty, whereas the HO head (right) produces a more uniform uncertainty profile.}
\label{fig:burgers_he_ho_decomp}
\vspace{-1.5em}
\end{figure*}

\begin{figure*}[t]
\centering
\scalebox{0.96}{%
\setlength{\tabcolsep}{2pt}

\begin{tabular}{ccccccc}
&
\scriptsize Input field &
\scriptsize Ground truth &
\scriptsize Predictive mean &
\scriptsize Epistemic std &
\scriptsize Aleatoric std &
\scriptsize Absolute error \\

% Darcy
\rotatebox{90}{\scriptsize \textbf{(a) Darcy flow}} &
\includegraphics[width=0.15\linewidth,trim=40 30 44 5,clip]{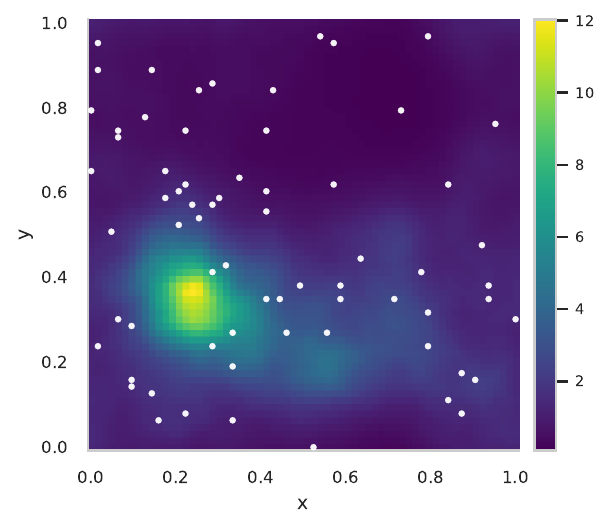} &
\includegraphics[width=0.15\linewidth,trim=40 30 50 5,clip]{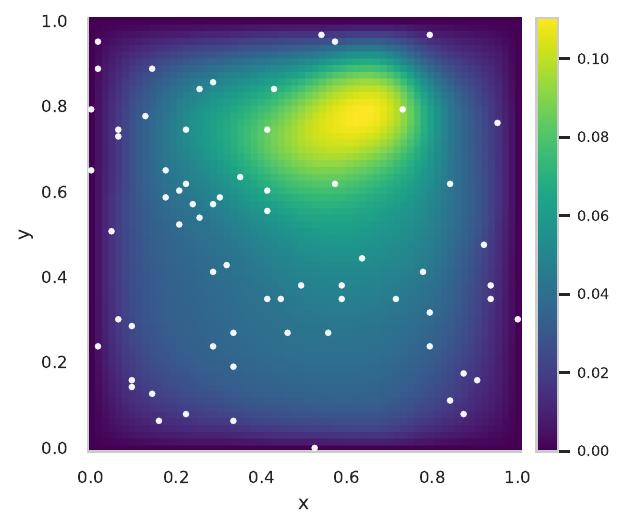} &
\includegraphics[width=0.15\linewidth,trim=40 30 50 5,clip]{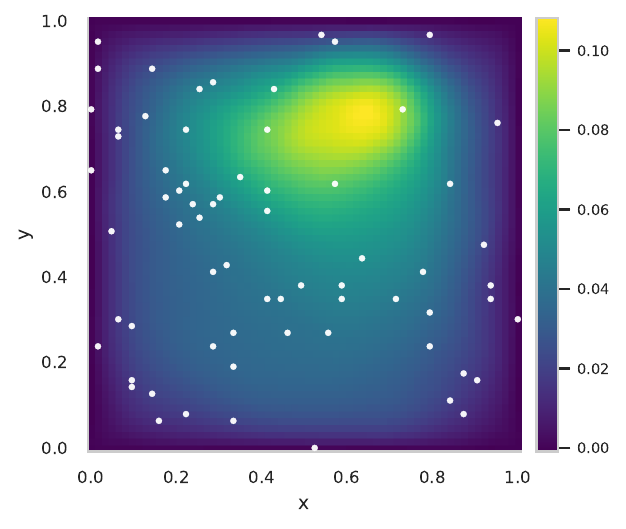} &
\includegraphics[width=0.15\linewidth,trim=40 30 58.8 5,clip]{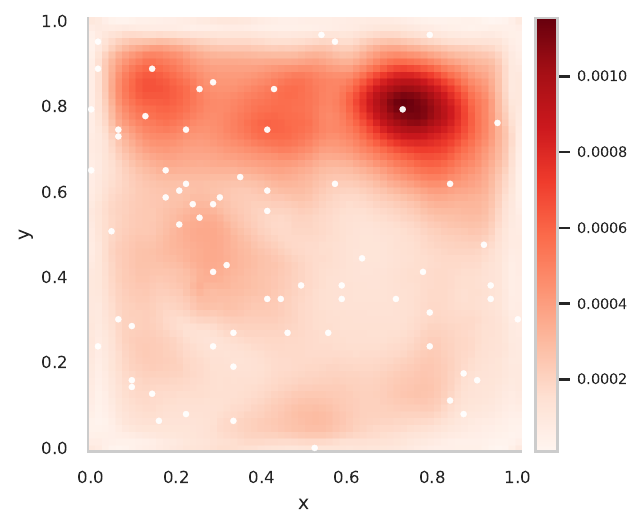} &
\includegraphics[width=0.15\linewidth,trim=40 30 58.7 5,clip]{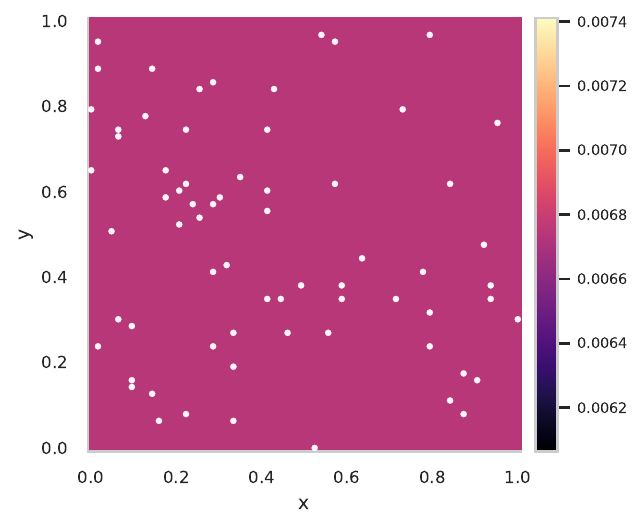} &
\includegraphics[width=0.15\linewidth,trim=40 30 54 5,clip]{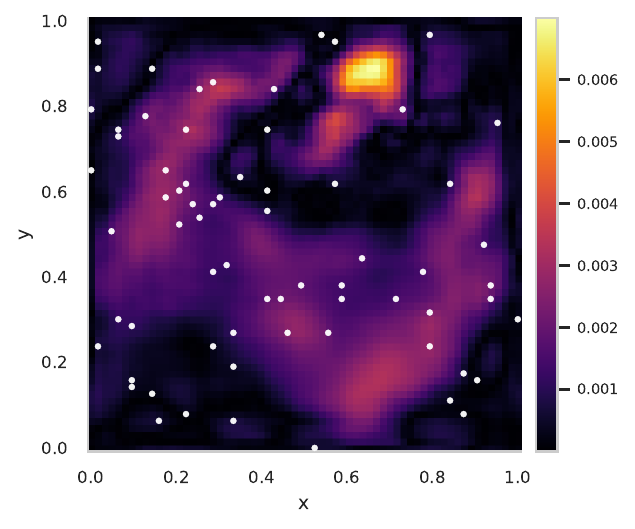} \\[0.1em]

% Navier-Stokes
\rotatebox{90}{\scriptsize \textbf{(b) NS}} &
\includegraphics[width=0.15\linewidth,trim=40 30 50 5,clip]{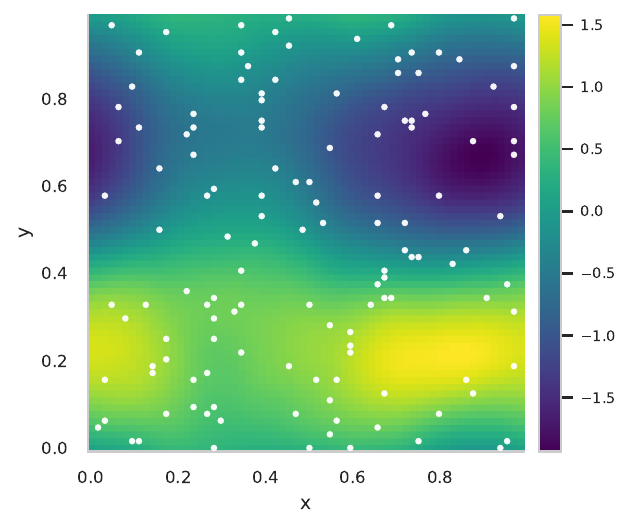} &
\includegraphics[width=0.15\linewidth,trim=40 30 50 5,clip]{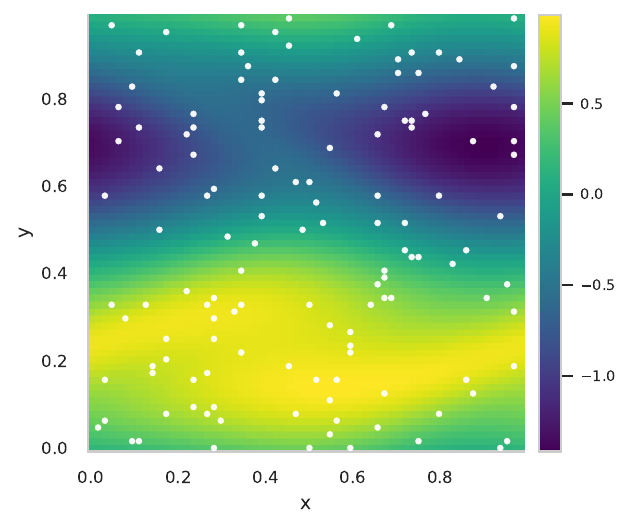} &
\includegraphics[width=0.15\linewidth,trim=40 30 50 5,clip]{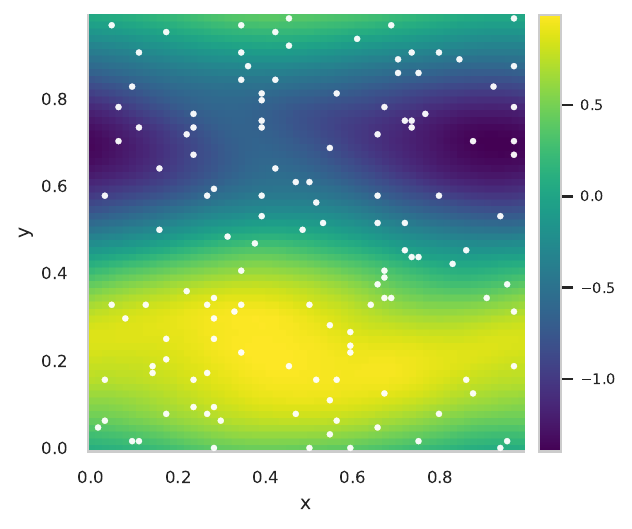} &
\includegraphics[width=0.15\linewidth,trim=40 30 50 5,clip]{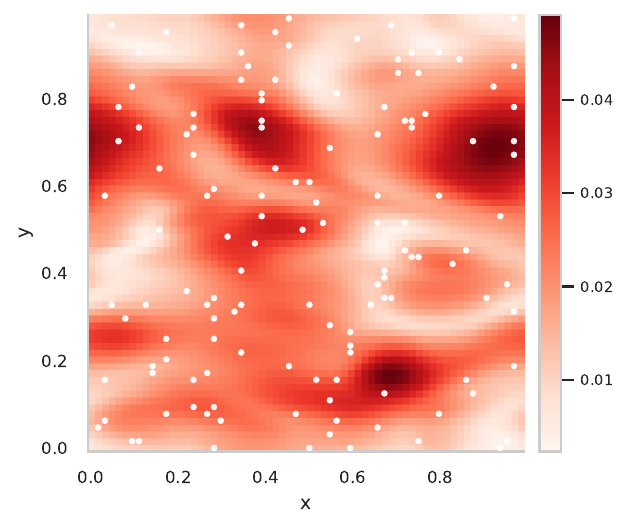} &
\includegraphics[width=0.15\linewidth,trim=40 30 54 5,clip]{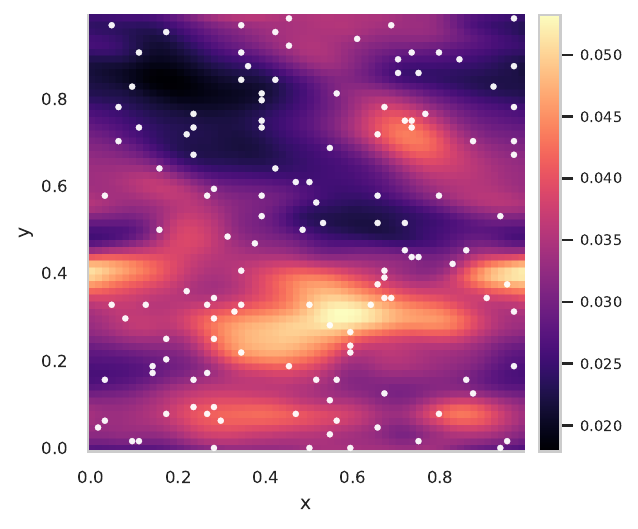} &
\includegraphics[width=0.15\linewidth,trim=40 30 50 5,clip]{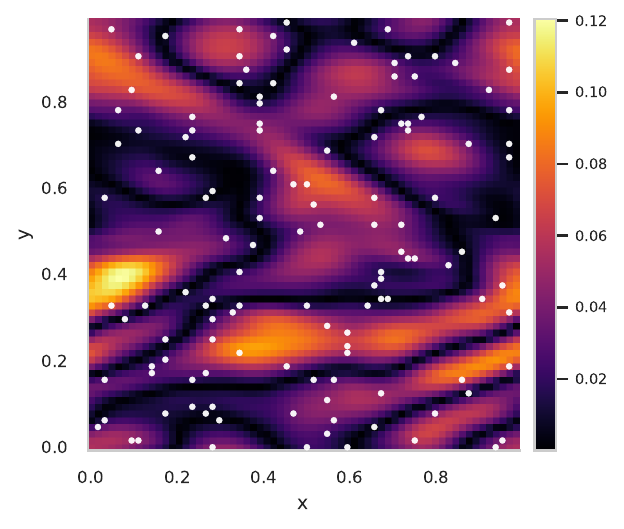} \\
\end{tabular}
}
\vspace{-0.5em}
\caption{\textbf{PANOP-HE behaviour on Darcy and Navier--Stokes.} The model separates epistemic and aleatoric uncertainty while maintaining accurate predictive means. On Darcy, uncertainty concentrates around the harder high-response region, whereas Navier--Stokes shows a broader, more structured pattern consistent with its more complex dynamics. White markers indicate context locations.}
\label{fig:darcy_navier_prob_he_qualitative}
\vspace{-1em}
\end{figure*}

\paragraph{Deterministic Sparse Operator Learning.} Table~\ref{tab:det_main} presents the main deterministic results across the three PDE benchmarks. Sparse conditional operator prediction is effective, both NOP variants achieve errors comparable to or even slightly lower than the standard dense-grid reference. Furthermore, explicitly preserving local context--query geometry via the attention pathway yields an often beneficial, but regime-dependent, reduction in relative $L^2$. For Burgers, where the dynamics are periodic and well aligned with the spectral decoder, attention improves deterministic performance even in the standard setting. In Navier--Stokes, by contrast, DNOP remains highly competitive under standard uniform evaluation, and the main advantage of attention appears instead in greater robustness under shifts in context sampling strategy and budget. More generally, attention becomes especially valuable when local spatial topology is irregular and task-critical, improving robustness to both the number of observed context points and their spatial distribution, and outperforming the convolution-only variant in these more challenging settings (see Appendix \ref{sec:appendix_context_ablations} and \ref{sec:train_eval_matrix}).

\paragraph{Probabilistic Sparse Operator Learning.} Table ~\ref{tab:prob_post_prior_combined} summarizes the base probabilistic results for both variance parameterizations, with prior-mean as the main test-time setting and posterior-mean as an oracle-style diagnostic of decoder quality when the latent is target-informed. The results show that latent stochastic conditioning preserves the predictive capacity of the architecture while providing meaningful uncertainty. Across all three PDE families, posterior and prior performance remain close for all models, indicating that the learned prior is reasonably aligned with the posterior and that test-time prior inference loses little information. The predictive distributions remain informative across regimes, although calibration quality is benchmark-dependent, with conservative coverage in some settings and mild under-coverage in Navier–Stokes. Zero-latent evaluation degrades substantially across all PDE benchmarks, and more for PNOP than PANOP, suggesting that geometry-aware conditioning better supports latent stochastic inference. Appendix~\ref{sec:probabilistic_ablations} reinforces this by showing that prior-sampled performance remains close to prior-mean for the strongest variants.

On Burgers, probabilistic performance remains close to the deterministic setting, with negligible posterior--prior gaps. Thus, uncertainty can be added with little effect on the predictive mean, making Burgers the closest PDE analogue to the GP diagnostic regime.

Across benchmarks, PANOP is generally the strongest probabilistic architecture. However, the preferred variance parameterization is benchmark-dependent. For Burgers, PANOP-HE is the strongest overall, while PANOP-HO attains higher nominal coverage at the cost of worse error, likelihood, and substantially wider intervals. Figure~\ref{fig:burgers_he_ho_decomp} depicts the corresponding uncertainty decomposition, showing that the HO head yields a comparatively more uniform uncertainty level, whereas the HE head is more expressive and introduces useful spatial adaptation near sharper transitions.

On Darcy, the variance head plays a particularly important role. PANOP-HO is the strongest probabilistic model under both posterior and prior evaluation. Given that the widths are similar across models, this improvement reflects a better predictive mean rather than trivial over-dispersion. This is consistent with broader ablations on Darcy (see Appendix \ref{sec:local_conditioning} and \ref{sec:probabilistic_ablations}), where the HO head appears to act as a useful regularizer by limiting the tendency of the model to absorb difficulty through variance inflation. Accordingly, Figure~\ref{fig:darcy_navier_prob_he_qualitative} shows that the predictive mean remains accurate, while uncertainty is localized in the harder regions of the field rather than becoming broadly diffuse.

Navier--Stokes is the most sensitive benchmark with respect to the variance head. Both architectures deteriorate sharply in the HO setting, indicating that a single global noise scale is too restrictive for this heterogeneous dynamical regime. By contrast, PANOP-HE is best under both posterior and prior evaluation, and uncertainty remains well behaved under prior-conditioned inference. While the HO variants produce substantially wider intervals, they remain clearly worse than their HE counterparts. Consistently, Figure~\ref{fig:darcy_navier_prob_he_qualitative} shows that uncertainty in Navier--Stokes is not only larger than in Darcy, but also more spatially structured, in line with the greater complexity of the underlying dynamics.

Similar to the deterministic setting, the role of observation geometry varies across operator regimes and is particularly important in the two-dimensional settings, where geometry-aware conditioning and training under more diverse observation patterns improve robustness. Likewise, cross-resolution transfer under sparse conditional inference remains limited overall (Appendix~\ref{sec:cross_resolution}). While Burgers exhibits meaningful zero-shot transfer, it is much more difficult in the harder sparse 2D regimes.

Overall, the learned priors are often well aligned with the posteriors across architectures, PANOP is often the strongest probabilistic model, and variance parameterization preference is task-dependent: Burgers and Navier-Stokes favor HE heads and Darcy favors the added regularization of a HO head.

\begin{table}[t]
\caption{Probabilistic models under posterior-mean and prior-mean evaluation. Mean values over 4 seeds. Lower rel-$L^2$, MC-NLL, and width are better; coverage is best near the 0.95 nominal level}
\label{tab:prob_post_prior_combined}
\centering
\small
\setlength{\tabcolsep}{4pt}
\resizebox{\linewidth}{!}{%
\begin{tabular}{ll@{\hspace{3pt}}cccc@{\hspace{6pt}}cccc}
\toprule
& & \multicolumn{4}{c}{\textbf{Posterior-mean}} & \multicolumn{4}{c}{\textbf{Prior-mean}} \\
\cmidrule(lr){3-6} \cmidrule(lr){7-10}
\textbf{PDE} & \textbf{Model} & \textbf{Rel-$L^2$} & \textbf{MC-NLL} & \textbf{Cov.} & \textbf{Width} & \textbf{Rel-$L^2$} & \textbf{MC-NLL} & \textbf{Cov.} & \textbf{Width} \\
\midrule
\multirow{4}{*}{Burgers}
& PNOP-HE  & $0.0096 \spm 0.0007$ & $-989$ & $0.971$ & $0.025$ & $0.0096 \spm 0.0006$ & $-989$ & $0.972$ & $0.025$ \\
& PNOP-HO  & $0.0098 \spm 0.0006$ & $-991$ & $0.971$ & $0.025$ & $0.0098 \spm 0.0005$ & $-991$ & $0.970$ & $0.025$ \\
& PANOP-HE & $0.0089 \spm 0.0002$ & $-1001$ & $0.974$ & $0.025$ & $0.0089 \spm 0.0003$ & $-1001$ & $0.974$ & $0.025$ \\
& PANOP-HO & $0.0110 \spm 0.0006$ & $-862$ & $0.988$ & $0.045$ & $0.0110 \spm 0.0006$ & $-862$ & $0.988$ & $0.045$ \\
\midrule
\multirow{4}{*}{Darcy}
& PNOP-HE  & $0.0727 \spm 0.0023$ & $-16627$ & $0.996$ & $0.024$ & $0.0730 \spm 0.0020$ & $-16036$ & $0.996$ & $0.024$ \\
& PNOP-HO  & $0.0704 \spm 0.0033$ & $-16627$ & $0.996$ & $0.024$ & $0.0703 \spm 0.0033$ & $-15754$ & $0.997$ & $0.024$ \\
& PANOP-HE & $0.0712 \spm 0.0036$ & $-16617$ & $0.996$ & $0.024$ & $0.0712 \spm 0.0036$ & $-16617$ & $0.996$ & $0.024$ \\
& PANOP-HO & $0.0657 \spm 0.0001$ & $-16672$ & $0.998$ & $0.024$ & $0.0657 \spm 0.0002$ & $-16672$ & $0.998$ & $0.024$ \\
\midrule
\multirow{4}{*}{NS}
& PNOP-HE  & $0.0511 \spm 0.0074$ & $-12865$ & $0.907$ & $0.062$ & $0.0543 \spm 0.0118$ & $-12392$ & $0.890$ & $0.062$ \\
& PNOP-HO  & $0.1246 \spm 0.0396$ & $-8352$ & $0.892$ & $0.108$ & $0.1258 \spm 0.0448$ & $-8058$ & $0.889$ & $0.108$ \\
& PANOP-HE & $0.0496 \spm 0.0048$ & $-13001$ & $0.903$ & $0.058$ & $0.0495 \spm 0.0105$ & $-12951$ & $0.902$ & $0.058$ \\
& PANOP-HO & $0.1180 \spm 0.0378$ & $-8508$ & $0.898$ & $0.107$ & $0.1185 \spm 0.0381$ & $-8499$ & $0.897$ & $0.107$ \\
\bottomrule
\end{tabular}
}
\vspace{-1.3em}
\end{table}

\section{Discussion and Further Work}
\label{sec:discussion}

Despite observing only a small fraction of the response field, \textsc{NOP}s attain comparable predictive performance to dense-grid baselines and in some cases improve upon it, showing that sparse conditional operator learning is viable across diverse PDE families. The framework successfully incorporates uncertainty quantification, and naturally supports its decomposition into a global latent and a local likelihood component. The latent captures task-level ambiguity induced by sparse observations, while the output head models local predictive uncertainty. The consistent small gap between prior-based and posterior-informed evaluations suggests that amortized inference can extract task-specific representations from sparse context that remain effective at test time.

At the same time, the preferred architectural configuration depends on both the regime and the objective. Smooth periodic regimes are often well served by coarse set-based conditioning, whereas more geometry-sensitive 2D regimes require stricter preservation of local geometry. More broadly, the preferred conditioning pathway also depends on whether the task is deterministic reconstruction or uncertainty-aware inference. Likewise, the preferred variance parameterization is regime-dependent: some settings benefit from the greater flexibility of heteroscedastic heads, whereas others, such as Darcy, benefit from the regularization induced by more constrained parameterizations.

\paragraph{Failure Modes and Design Principles.}
The success of sparse conditional operator learning strongly depends on how geometric information is propagated. A central failure mode is the loss of spatial structure when local context is overly compressed into global summaries or strongly modulated by a global latent. This is relatively benign in smooth periodic settings such as Burgers, but much more damaging in non-periodic, boundary-sensitive regimes such as Darcy, where it leads to diffuse predictive means and compensatory uncertainty. The results suggest that the main bottleneck is not simply model capacity, but the stability of the conditioning interface between sparse context, latent modulation, and operator decoding. In practice, this makes geometry-sensitive regimes particularly vulnerable to architectural choices that blur local structure or allow the uncertainty mechanism to absorb prediction difficulty too easily. These are distinct bottlenecks and should be addressed separately. First, local geometry should be preserved explicitly whenever the regime is geometry-sensitive. Second, latent mechanisms and variance heads should complement the spatial inductive biases carried by the deterministic trunk.

\paragraph{Limitations and future work.}
Our benchmarks do not cover strongly chaotic dynamics, highly multiscale systems, or irregular geometries. The probabilistic models use a single global Gaussian latent and diagonal likelihoods, which limits the representation of localized epistemic uncertainty and explicit spatial covariance. A trade-off between deterministic accuracy and uncertainty-aware prediction appears, especially in the harder 2D regimes where robust sparse cross-resolution transfer remains challenging. Probabilistic extensions increase training-time memory and runtime cost, although inference remains modest in our profiling (Appendix~\ref{sec:cost_complexity}). Since the decoder backbone is fixed to an FNO architecture, observed failure modes, e.g., geometry sensitivity in Darcy flow and stronger cross-resolution degradation in 2D, may reflect the interaction between sparse conditional representations and a spectral decoder biased toward regular periodic structure, rather than a core limitation of the NOP framework itself. These findings motivate future work on spatially structured latent representations, richer likelihoods with spatial covariance, and decoder families better suited to non-periodic boundaries and irregular geometries, extending our architectures to richer properties and a broader range of applications.

\section{Conclusion}
\label{sec:conclusion}

We introduce Neural Operator Processes (\textsc{NOP}s), a framework for operator learning under partial observations that unifies the set-conditioned inference of NPs with the continuous-space mappings of NOs. Across different benchmarks, our evaluation shows that sparse conditional operator prediction is feasible and can perform competitively with, or even exceed, standard dense-grid baselines. Concretely, NOPs match or surpass dense-grid FNO using only 0.78--25\% of grid points, suggesting savings in sensing or storage when dense measurements are costly. Crucially, the probabilistic extensions of \textsc{NOP}s augment this predictive strength with coherent uncertainty estimates for full-field predictions. Beyond empirical performance, this work clarifies the structural principles that govern uncertainty-aware operator learning. We demonstrated that explicitly preserving local geometry is a key design requirement, although its importance varies across benchmarks. Furthermore, we showed that reliable probabilistic inference requires latent conditioning and variance parameterizations that complement, rather than overwrite, the spatial inductive biases of the deterministic trunk. Overall, \textsc{NOP}s offer a practical, explicit, and extensible framework for combining sparse conditional inference, structured operator decoding, and uncertainty quantification, and clarify the architectural principles that govern success in this setting.

%%%%%%%%%%%%%%%%%%%%%%%%%%%%%%%%%%%%%%%%%%%%%%%%%%%%%%%%%%%%
\newpage

\bibliographystyle{plainnat}
\bibliography{nops_conferences}

\newpage
\appendix

%%%%%%%%%%%%%%%%%%%%%%%%%%%%%%%%%%%%%%%%%%%%%%%%%%%%%%%%%%%%

\section{Overview of Extended Analysis and Ablations}
\label{sec:appendix_overview}

This appendix complements the main text with a broader set of diagnostics, robustness analyses, and practical profiling results. Its purpose is twofold: first, to clarify how the probabilistic latent mechanisms behave beyond the main summary tables; and second, to stress-test the NOP framework under changes in observation geometry, context availability, discretization, architectural sub-components, and computational budget. Taken together, these studies are intended to distinguish which empirical gains arise from the general NOP formulation and which depend more specifically on particular conditioning mechanisms or evaluation regimes. Unless otherwise stated, results are reported as means over four independent seeds. Some appendix ablations use separately trained checkpoint suites, so matching labels across tables do not always denote repeated evaluations of the same checkpoint. Standard deviations are shown for the primary comparison metrics, especially relative-$L^2$ errors and the main quantities supporting model comparisons. In wider diagnostic tables with many columns, some secondary metrics are reported as means only to preserve readability; these entries are computed from the same set of runs.

For clarity, the appendix is organized into two parts:

\paragraph{Probabilistic diagnostics, auxiliary baselines, and latent analyses (Section \ref{app:prob_latent_diagnostics}).}
We begin with extended GP diagnostics, latent robustness tests, and a latent injection ablation. These diagnostics clarify how posterior-based, prior-based, and latent-ablated evaluations should be interpreted, and support the probabilistic design choices beyond the condensed comparisons in the main text. In particular, they separate decoder quality under target-informed latent inference from deployment-relevant prior behavior, and show why stable prior-conditioned inference is more informative than posterior reconstruction alone. We also include a supplementary deterministic comparison with additional DeepONet-based neural operators, considered here as classical branch--trunk baselines, to position NOP relative to another canonical non-spectral operator-learning family with architectural properties different from those of FNO.

\paragraph{Extended architectural and generalization ablations (Section \ref{app:arch_diagnostics}).}
The second part collects the broader ablation suite on observation geometry, test-time context scaling, train--evaluation strategy mismatch, cross-resolution transfer, local conditioning decomposition, and computational cost. The sequence is structured to move from data-distribution sensitivity, to out-of-distribution robustness, to mechanistic architectural dissection, and finally to practical cost profiling.

\paragraph{Context strategy and context budget ablations (Section \ref{sec:appendix_context_ablations}).}
We first establish the in-distribution sensitivity of the models to the geometry and cardinality of the observation set. This provides the empirical baseline for the appendix: before testing transfer or robustness, we characterize how strongly sparse conditional operator learning depends on where observations are placed and how many are available.

\paragraph{Test-time context scaling (Section \ref{sec:test_time_scaling}).}
We then test whether the limitations identified in the previous section can be alleviated simply by supplying more context at inference time while keeping the checkpoint fixed. This serves as a direct control for the context-budget study, separating representational limitations from the trivial effect of additional observations at deployment.

\paragraph{Train--evaluation strategy ablation (Section \ref{sec:train_eval_matrix}).}
Having characterized in-distribution behavior, we next move to out-of-distribution geometry shifts by evaluating checkpoints trained under one observation strategy and tested under another. This section closes the geometry-focused part of the appendix by showing when robustness comes from architecture, when it comes from training distribution, and how much specialization to sensor layout remains.

\paragraph{Cross-resolution transfer (Section \ref{sec:cross_resolution}).}
We then shift from observation-space variation to discretization-space variation, testing whether coarse-trained models transfer zero-shot to finer grids. This section connects the sparse conditional setting back to a typical objective in operator learning, namely resolution transfer, while also making explicit where discretization invariance remains limited under partial observations.

\paragraph{Local conditioning decomposition (Section \ref{sec:local_conditioning}).}
After documenting the main robustness and transfer phenomena, we dissect the local conditioning trunk itself by isolating the SetConv and query-aligned attention pathways. This provides a more mechanistic view of which component contributes most strongly under different PDE families and inference objectives, and helps justify the combined design used in the main-text models.

\paragraph{Computational cost and complexity (Section \ref{sec:cost_complexity}).}
Finally, we profile parameter count, memory footprint, and training/inference cost. This concludes the appendix with a practical perspective, showing the computational trade-offs associated with the geometry-aware and probabilistic extensions.

\paragraph{Scope and architectural limitation.}
One important scope condition applies across all of these studies: throughout the paper and appendix, we keep the neural operator decoder backbone fixed to an FNO. This allows the ablations to isolate the effects of conditional representation, latent modeling, and geometry-aware conditioning without conflating them with backbone changes. At the same time, it also limits the scope of the conclusions. In particular, some of the weaknesses observed in the more difficult transfer settings, especially in non-periodic or multi-scale 2D regimes, may reflect not only limitations of the sparse conditional representation but also the interaction between that representation and the spectral FNO decoder. The appendix therefore should not be read as claiming that all remaining failure modes are intrinsic to the NOP framework itself. Rather, the results establish that the NOP framework yields clear gains in robustness, probabilistic stability, and geometry-aware conditioning under a fixed and widely used operator-decoder family, while also identifying decoder-sensitive regimes that remain promising directions for future extensions. Task-matched comparisons with probabilistic neural-operator baselines remain future work, as many require adapting dense-input uncertainty estimators to sparse response-conditioned inference.

\subsection{Reproducibility and Ablation Setup}

\paragraph{Compute Environment}
All models and dataloaders are implemented in Python~3.11.7 using \href{https://github.com/pytorch/pytorch}{PyTorch}~2.5.1 with CUDA~12.7 \citep{paszke_pytorch_2019}. Experiment configuration is managed with \href{https://hydra.cc}{Hydra} and \href{https://github.com/omry/omegaconf}{OmegaConf}\citep{yadan2019hydra}. The neural-operator decoder uses an FNO backbone based on the Fourier Neural Operator architecture of \citet{li2021fno}, following the publicly available \href{https://github.com/zongyi-li/fourier_neural_operator}{Fourier Neural Operator implementation}. Experiments were run on a single NVIDIA GH200 GPU node with 96\,GB of GPU memory, an ARM Neoverse-V2 CPU with 72~cores (up to 3.47\,GHz), and 477\,GB of system RAM, under Linux (aarch64). Computational profiling results are reported in Section~\ref{sec:cost_complexity}. All datasets were generated by the authors. External software dependencies and referenced public codebases are used under their respective licenses; the primary assets include PyTorch (BSD-3-Clause), Hydra (MIT), OmegaConf (BSD-3-Clause), and the referenced FNO codebase (MIT).

\paragraph{Context Budgets.} For some ablations, models are evaluated across three context budget regimes: low, base, and high. The absolute number of context points defining these regimes varies by benchmark to account for differences in domain dimensionality. Specifically, the low, base, and high budgets correspond to 8--16, 16--64, and 64--128 points for Burgers 1D; 16--32, 32--256, and 256--512 points for Darcy 2D; and 32--64, 64--256, and 256--512 points for Navier--Stokes 2D.

\paragraph{Context sampling strategies.}
Uniform contexts sample indices uniformly without replacement. Clustered contexts draw a random center in the domain and select the $n_C$ nearest grid points, using normalized absolute distance in 1D and normalized elliptical distance in 2D, with each coordinate normalized by $20\%$ of its domain range. Mixed contexts use clustered sampling with probability $0.5$ and uniform sampling otherwise.

\paragraph{Model Configuration.}
Model configurations are kept largely consistent across experiments, using hidden dimension~64, two-layer input/output projections, four FNO layers, coordinate features, and forward FFT normalization. Fourier modes are set to 24 in 1D and 16 per spatial dimension in 2D. Probabilistic variants use a 16-dimensional latent variable, two-layer latent encoders with hidden dimension~128, linear $\beta$-warmup to $1$ over 50 epochs, and 8 test-time Monte Carlo samples; attention-based variants use four attention heads. Dataset-specific changes are mainly limited to boundary features and SetConv lengthscales. Additional configuration and hyperparameter details can be found in the accompanying code.

\section{Probabilistic diagnostics and latent analyses}
\label{app:prob_latent_diagnostics}

\begin{table}[t]
\centering
\setlength{\tabcolsep}{2pt}
\caption{Homoscedastic variants rel-$L^2$ on GP benchmark with different latent inference modes. Mean values over 4 seeds. Lower is better.}
\label{tab:gp_combined_ho}
\resizebox{0.85\linewidth}{!}{%
\begin{tabular}{lcccccc}
\toprule
& \multicolumn{2}{c}{\textbf{Prior-mean}} & \multicolumn{2}{c}{\textbf{Posterior-mean}} & \multicolumn{2}{c}{\textbf{Zero-latent}} \\
\cmidrule(lr){2-3} \cmidrule(lr){4-5} \cmidrule(lr){6-7}
\textbf{GP} & \textbf{PNOP-HO} & \textbf{PANOP-HO} & \textbf{PNOP-HO} & \textbf{PANOP-HO} & \textbf{PNOP-HO} & \textbf{PANOP-HO} \\
\midrule
RBF 
& $0.301 \spm 0.001$ & $0.331 \spm 0.026$
& $0.221 \spm 0.001$ & \textbf{$0.145 \spm 0.064$} 
& $0.735 \spm 0.016$ & $1.014 \spm 0.230$ \\
Mixed 
& $0.393 \spm 0.001$ & $0.427 \spm 0.015$
& $0.322 \spm 0.000$ & \textbf{$0.225 \spm 0.044$} 
& $0.769 \spm 0.013$ & $1.041 \spm 0.145$ \\
Mat\'ern 
& $0.444 \spm 0.001$ & $0.465 \spm 0.019$
& $0.375 \spm 0.001$ & \textbf{$0.285 \spm 0.078$} 
& $0.745 \spm 0.007$ & $0.977 \spm 0.193$ \\
\bottomrule
\end{tabular}
}
\end{table}

\subsection{Additional GP Diagnostics}
\label{app:gp}

We provide additional GP diagnostics to disentangle the deterministic sparse reconstruction quality, posterior-informed decoder quality, and deployment-relevant prior-based probabilistic behavior.

\paragraph{Latent behavior and variance parameterization.}
As shown in Table~\ref{tab:gp_combined_ho}, the HO variants follow the same basic pattern in the GPs as the HE results in Table~\ref{tab:gp_combined_main}. Posterior-mean evaluation is much stronger than zero-latent evaluation, confirming that the latent pathway is active, while prior-mean remains reasonably close to posterior-mean in the stable regimes. At the same time, PANOP-HO configuration exhibits a larger prior--posterior gap and substantially higher variability across seeds than the HE variants in Table~\ref{tab:gp_combined_main}, despite occasionally very low posterior-mean errors.

\begin{figure*}[t]
\centering
\begin{minipage}{0.5\linewidth}
\centering
\includegraphics[width=\linewidth]{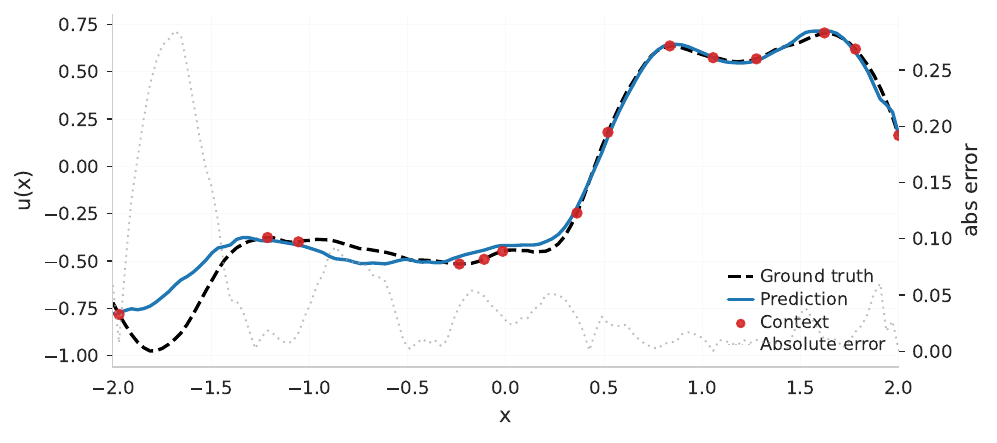}
\end{minipage}\hfill
\begin{minipage}{0.5\linewidth}
\centering
\includegraphics[width=\linewidth]{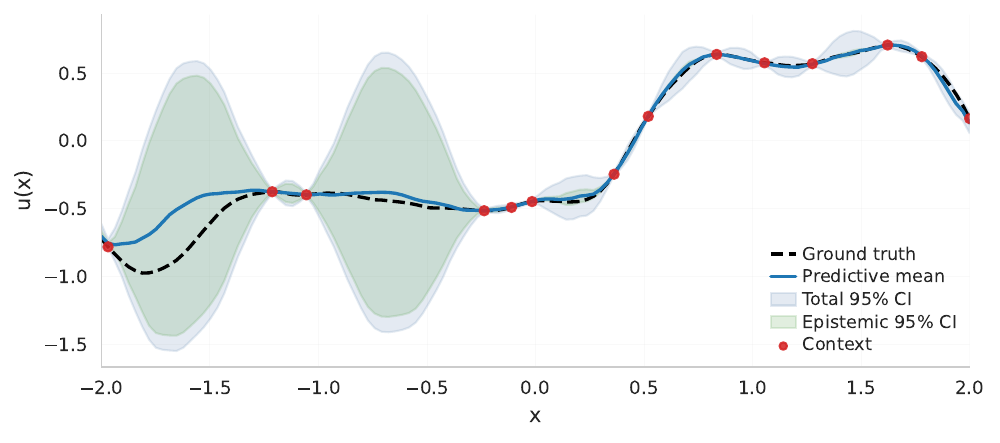}
\end{minipage}
\caption{\textbf{Deterministic NOP and probabilistic PNOP behavior on 1D GP-Matérn}. Left: deterministic prediction from sparse context observations, with larger errors across wider context gaps and sharper local variations. Right: probabilistic prediction with a HE head. The predictive mean remains accurate near context points, while uncertainty expands in underconstrained intervals. The epistemic band is computed from variance across latent samples from the learned conditional prior $p_\theta(z\mid C)$, and the total band additionally includes the predicted conditional likelihood variance.}
\label{fig:prob_1d_gp_matern}
\vspace{-1em}
\end{figure*}

\begin{figure*}[t]
\centering
\begin{minipage}{0.5\linewidth}
\centering
\includegraphics[width=\linewidth]{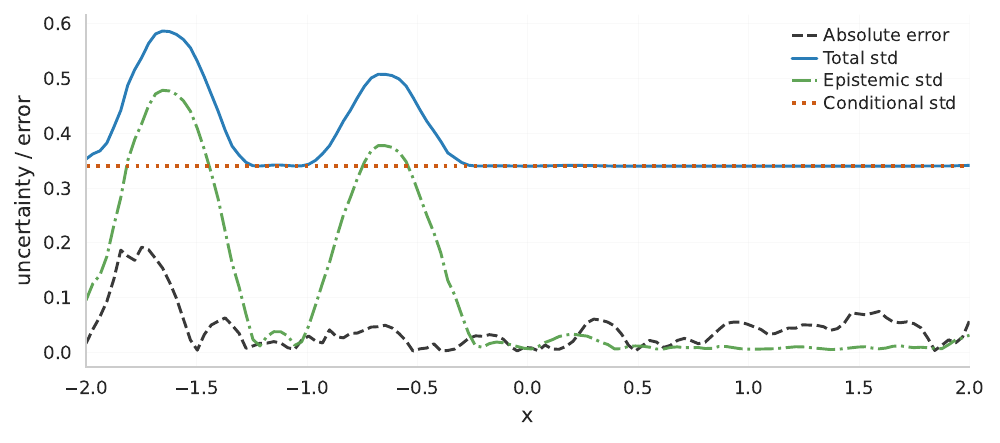}
\end{minipage}\hfill
\begin{minipage}{0.5\linewidth}
\centering
\includegraphics[width=\linewidth]{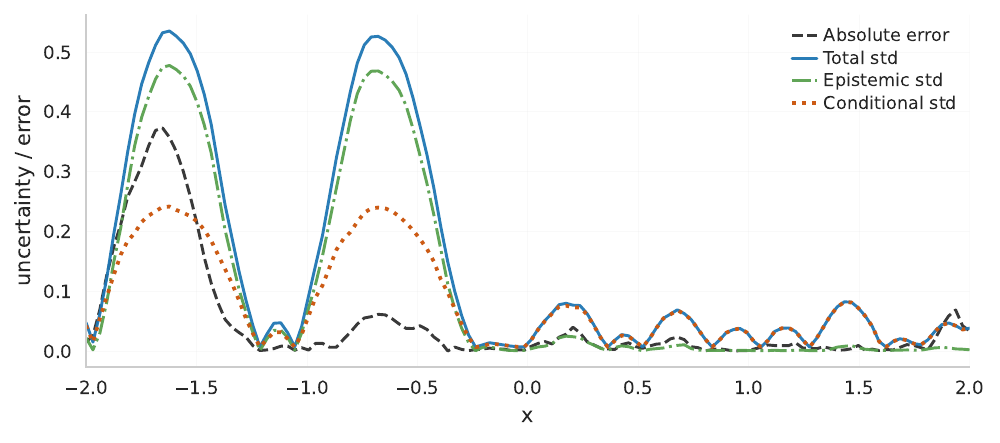}
\end{minipage}
\caption{\textbf{PNOP variance decomposition for the GP-Matérn example in Figure~\ref{fig:prob_1d_gp_matern}.} In the HO case (left), the conditional standard deviation remains nearly constant and the total uncertainty is driven mainly by the epistemic component. In the HE case (right), both epistemic and conditional terms vary across the domain, yielding a more spatially adaptive uncertainty profile. In both parameterizations, uncertainty is largest in the least constrained regions.}
\label{fig:prob_1d_gp_matern_decomp}
\end{figure*}

Figure~\ref{fig:prob_1d_gp_matern} illustrates the transition from deterministic to probabilistic prediction on GP-Mat\'ern. DNOP reconstructs the function well near observed regions, while the error increases across longer context gaps and around sharper local variations. The probabilistic extension (PNOP) preserves a similar predictive mean in well constrained regions, but in sparsely observed intervals the mean becomes smoother than the ground truth and the predictive uncertainty expands substantially. Figure~\ref{fig:prob_1d_gp_matern_decomp} shows that with an HO head the conditional term remains nearly constant and the spatial variation of total uncertainty is driven mainly by the epistemic component, yielding a cleaner and more regular uncertainty profile. With an HE head, the uncertainty becomes more spatially adaptive, with a larger share of predictive spread assigned to the conditional variance term. In both cases, uncertainty is largest in the least constrained regions, and the Mat\'ern example makes visible that the main source of ambiguity comes from sparse conditioning, while the choice of variance head changes how that ambiguity is distributed across the domain.

\paragraph{Posterior-informed decoder quality.}
Considering Tables~\ref{tab:gp_combined_main} and~\ref{tab:gp_combined_ho}, all probabilistic variants improve the posterior-informed decoder quality over the deterministic baselines in this regime, with gains that are most pronounced on RBF and more moderate on Matérn-5/2. Once the latent is well informed by the target, the decoder has sufficient capacity to reconstruct the function accurately. At the same time, PANOP-HO remains stronger than PNOP-HO, reflecting not only latent inference quality and robustness but also how effectively each architecture supports the interaction between latent conditioning and decoder reconstruction.

\paragraph{Prior-based predictive behavior.}
Table~\ref{tab:gp_prior_sample_app} reports predictive quality and uncertainty metrics under prior sampling. Across the stable variants, relative-$L^2$ and interval width increase from RBF to Mixed to Mat\'ern, while coverage remains broadly near nominal. This is consistent with the increasing roughness of the task distribution. The learned priors remain well behaved at test time and capture task-level uncertainty in a way that yields meaningful stochastic predictions rather than pathological or trivially over-dispersed behavior.

\begin{table*}[t]
\centering
\caption{GP prior-sampled predictive performance. Lower is better for rel-$L^2$, MC-NLL, and width. Coverage is best when close to nominal 0.95 level. Mean values over 4 seeds.}
\label{tab:gp_prior_sample_app}
\resizebox{0.8\linewidth}{!}{%
\begin{tabular}{llcccc}
\toprule
\textbf{GP} & \textbf{Model} & \textbf{Rel-$L^2$} & \textbf{MC-NLL} & \textbf{Coverage} & \textbf{Avg.\ width} \\
\midrule
\multirow{4}{*}{RBF}
& PNOP-HE  & $0.278 \spm 0.002$ & $-238.02$ & $0.920$ & $0.699$ \\
& PNOP-HO  & $0.301 \spm 0.001$ & $-237.91$ & $0.920$ & $0.698$ \\
& PANOP-HE & $0.283 \spm 0.004$ & $-237.25$ & $0.921$ & $0.709$ \\
& PANOP-HO & $0.366 \spm 0.047$ & $460.86$  & $0.931$ & $1.070$ \\
\midrule
\multirow{4}{*}{Mixed}
& PNOP-HE  & $0.375 \spm 0.002$ & $-130.50$ & $0.897$ & $0.966$ \\
& PNOP-HO  & $0.393 \spm 0.001$ & $-130.44$ & $0.902$ & $0.983$ \\
& PANOP-HE & $0.379 \spm 0.002$ & $-130.65$ & $0.902$ & $0.993$ \\
& PANOP-HO & $0.457 \spm 0.022$ & $183.37$  & $0.920$ & $1.391$ \\
\midrule
\multirow{4}{*}{Mat\'ern}
& PNOP-HE  & $0.432 \spm 0.002$ & $-80.42$ & $0.899$ & $1.162$ \\
& PNOP-HO  & $0.444 \spm 0.001$ & $-79.98$ & $0.896$ & $1.154$ \\
& PANOP-HE & $0.433 \spm 0.002$ & $-81.27$ & $0.899$ & $1.165$ \\
& PANOP-HO & $0.491 \spm 0.033$ & $141.78$ & $0.913$ & $1.527$ \\
\bottomrule
\end{tabular}
}
\end{table*}

\subsection{Latent Probabilistic Robustness}
\label{sec:probabilistic_ablations}

\begin{table}[t]
\caption{Probabilistic models under prior-sample evaluation. Mean values over 4 seeds. Lower rel-$L^2$ and MC-NLL are better.}
\label{tab:prob_prior_sample}
\centering
\setlength{\tabcolsep}{4pt}
\resizebox{0.73\linewidth}{!}{%
\begin{tabular}{llcccc}
\toprule
\textbf{Benchmark} & \textbf{Model} & \textbf{Rel-$L^2$} & \textbf{MC-NLL} & \textbf{Coverage} & \textbf{Width} \\
\midrule
\multirow{4}{*}{Burgers}
& PNOP-HE  & $0.0097 \spm 0.0007$ & $-997.4$  & $0.971$ & $0.025$ \\
& PNOP-HO  & $0.0403 \spm 0.0522$ & $-995.0$  & $0.974$ & $0.166$ \\
& PANOP-HE & $0.0088 \spm 0.0002$ & $-1002.3$ & $0.974$ & $0.025$ \\
& PANOP-HO & $0.0110 \spm 0.0005$ & $-861.8$  & $0.988$ & $0.045$ \\
\midrule
\multirow{4}{*}{Darcy}
& PNOP-HE  & $0.0732 \spm 0.0018$ & $-16606.2$ & $0.996$ & $0.024$ \\
& PNOP-HO  & $0.0705 \spm 0.0033$ & $-16627.3$ & $0.997$ & $0.024$ \\
& PANOP-HE & $0.0712 \spm 0.0037$ & $-16619.3$ & $0.996$ & $0.024$ \\
& PANOP-HO & $0.0657 \spm 0.0001$ & $-16671.9$ & $0.998$ & $0.024$ \\
\midrule
\multirow{4}{*}{NS}
& PNOP-HE  & $0.0551 \spm 0.0028$ & $-12639.7$ & $0.904$ & $0.066$ \\
& PNOP-HO  & $0.1260 \spm 0.0122$ & $-8034.1$  & $0.894$ & $0.110$ \\
& PANOP-HE & $0.0496 \spm 0.0024$ & $-12800.9$ & $0.902$ & $0.058$ \\
& PANOP-HO & $0.1187 \spm 0.0157$ & $-8261.7$  & $0.899$ & $0.107$ \\
\bottomrule
\end{tabular}
}
\vspace{-1em}
\end{table}

\begin{table}[t]
\caption{Probabilistic models under zero-latent evaluation. Mean values over 4 seeds. This is a stress test of latent dependence rather than a standard inference mode. Low rel-$L^2$ and MC-NLL are better.}
\label{tab:prob_zero_latent}
\centering
\resizebox{0.73\linewidth}{!}{%
\setlength{\tabcolsep}{4pt}
\begin{tabular}{llcccc}
\toprule
\textbf{Benchmark} & \textbf{Model} & \textbf{Rel-$L^2$} & \textbf{MC-NLL} & \textbf{Coverage} & \textbf{Width} \\
\midrule
\multirow{4}{*}{Burgers}
& PNOP-HE  & $0.2158 \spm 0.0237$ & $37221.0$  & $0.101$ & $0.025$ \\
& PNOP-HO  & $0.1913 \spm 0.0767$ & $26157.3$  & $0.142$ & $0.027$ \\
& PANOP-HE & $0.0194 \spm 0.0166$ & $-686.2$   & $0.860$ & $0.025$ \\
& PANOP-HO & $0.0226 \spm 0.0120$ & $-701.0$   & $0.915$ & $0.045$ \\
\midrule
\multirow{4}{*}{Darcy}
& PNOP-HE  & $0.1268 \spm 0.0342$ & $-15764.2$ & $0.949$ & $0.024$ \\
& PNOP-HO  & $0.1077 \spm 0.0298$ & $-16086.9$ & $0.968$ & $0.024$ \\
& PANOP-HE & $0.0865 \spm 0.0195$ & $-16424.3$ & $0.987$ & $0.024$ \\
& PANOP-HO & $0.0785 \spm 0.0108$ & $-16532.4$ & $0.993$ & $0.024$ \\
\midrule
\multirow{4}{*}{NS}
& PNOP-HE  & $0.4459 \spm 0.0416$ & $252007.2$ & $0.191$ & $0.051$ \\
& PNOP-HO  & $0.5350 \spm 0.0643$ & $115069.3$ & $0.409$ & $0.108$ \\
& PANOP-HE & $0.1900 \spm 0.0812$ & $53719.4$  & $0.400$ & $0.048$ \\
& PANOP-HO & $0.2696 \spm 0.0888$ & $18053.3$  & $0.632$ & $0.107$ \\
\bottomrule
\end{tabular}
}
\vspace{-1em}
\end{table}

\begin{figure}[t]
    \centering
    % Primera subfigura
    \begin{subfigure}{0.48\textwidth}
        \centering
        \includegraphics[width=\linewidth]{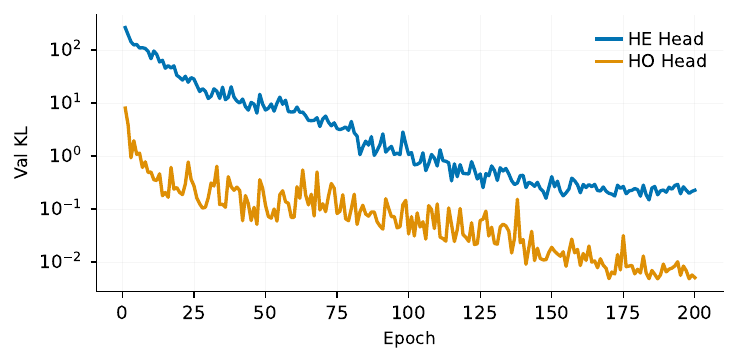}
        \caption{KL Divergence}
        \label{fig:results_kl}
    \end{subfigure}\hfill
    % Segunda subfigura
    \begin{subfigure}{0.48\textwidth}
        \centering
        \includegraphics[width=\linewidth]{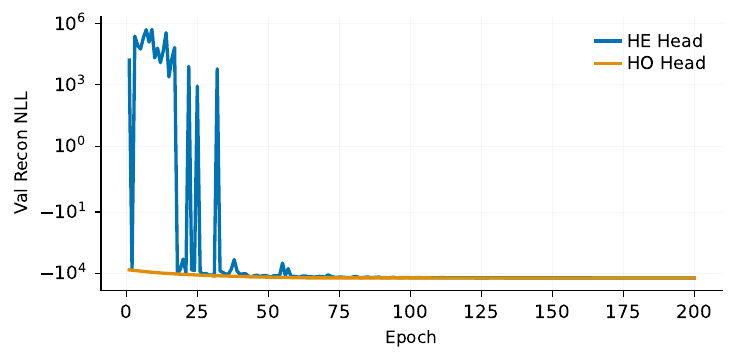}
        \caption{Reconstruction NLL}
        \label{fig:results_recon_nll}
    \end{subfigure}
    
    \caption{\textbf{Training dynamics of PANOP on Darcy.} The homoscedastic model maintains stable low-KL training; the heteroscedastic model exhibits early instability.}
    \label{fig:results_v2ba_training}
\end{figure}

Tables~\ref{tab:prob_prior_sample} and \ref{tab:prob_zero_latent} report two complementary probabilistic robustness analyses. Prior-sample evaluation probes the stochastic behavior of the learned prior more directly than prior-mean, while zero-latent evaluation acts as a stress test for how much useful information is carried by the latent pathway.

The prior-sample results in Table~\ref{tab:prob_prior_sample} broadly reinforce the conclusions from the main probabilistic results. In particular, PANOP remains very stable under stochastic prior draws. Across all PDE benchmarks, PANOP-HE and PANOP-HO are essentially unchanged relative to prior-mean evaluation, indicating that the learned prior is not only well aligned on average, but also stable under sampling. This is especially clear for PANOP-HE, whose prior-sample results remain nearly identical to its posterior-mean and prior-mean performance.

By contrast, PNOP shows mild to moderate sensitivity to prior sampling, which is mostly visible in Burgers under the HO parameterization. There, PNOP-HO exhibits a noticeable degradation together with a large increase in interval width and seed variability. Since the same configuration is much more stable under posterior-mean and prior-mean evaluation, this suggests that the issue is not simply poor mean prediction, but rather instability in the stochastic behavior of the learned prior under sampling. Outside of this case, PNOP remains reasonably stable, but still less consistent than PANOP.

The zero-latent results in Table~\ref{tab:prob_zero_latent} show that all models degrade, often sharply, confirming that the latent pathway carries meaningful task-specific information. On Burgers, removing the latent causes a severe collapse for PNOP, whereas PANOP degrades much less and remains far more accurate. The same qualitative trend appears on Navier--Stokes, where PANOP again deteriorates less severely than PNOP, especially in the HE setting. Darcy is less extreme, but the same ordering remains visible. Zero-latent performance is worse than posterior- and prior-based inference for all models, yet PANOP still retains a clear advantage over PNOP. Thus, the architecture is not merely achieving stronger results through a fragile latent dependence, but rather combining a more robust deterministic structure with a better behaved latent mechanism.

Furthermore, the robustness of the latent pathway is tightly coupled to the choice of variance parameterization. The training dynamics on Darcy in Figure~\ref{fig:results_v2ba_training} illustrates how the HO model enters a low-KL regime quickly with smooth optimization, effectively regularizing the latent space. In contrast, the HE model shows erratic early-phase KL spikes before stabilizing, reflecting the optimization difficulty of jointly learning a flexible local variance and a global latent prior in geometry-sensitive regimes. This reinforces why expressive HE heads require a highly stable deterministic trunk, such as PANOP, to prevent early training instabilities from permanently degrading the final learned prior.

These analyses support two additional conclusions. First, PANOP exhibits more stable prior behavior under sampling, which strengthens the interpretation that its prior is genuinely better aligned with test-time inference rather than only better in expectation. Second, although all models rely on latent conditioning to some extent, PANOP degrades more gracefully when the latent is removed, reflecting a more stable interaction between latent information and the underlying NO decoder.

\subsection{Latent Injection Mechanism Ablation}
\label{app:latent_injection_ablation}

\begin{table*}[t]
\caption{Latent injection ablation on Darcy for the probabilistic models. Mean values over 4 seeds. PNOP-m injects the global latent through decoder-wide FiLM modulation at each FNO block, whereas PNOP injects the latent once at the input stage before operator propagation.}
\label{tab:darcy_latent_injection_ablation}
\centering
\resizebox{0.8\linewidth}{!}{%
\setlength{\tabcolsep}{4pt}
\begin{tabular}{llcccc}
\toprule
\textbf{Model} & \textbf{Eval Mode} & \textbf{Rel-$L^2$} & \textbf{MC-NLL} & \textbf{Coverage} & \textbf{Width} \\
\midrule
\multirow{4}{*}{PNOP-m-HE}
& Posterior-mean & $0.0677 \spm 0.0006$ & $-16652.3$ & $0.997$ & $0.024$ \\
& Prior-mean     & $0.1295 \spm 0.0277$ & $-15472.4$ & $0.934$ & $0.025$ \\
& Prior-sample   & $0.1571 \spm 0.0162$ & $-16571.7$ & $0.989$ & $0.034$ \\
& Zero-latent    & $1.1086 \spm 0.4235$ & $-6456.0$  & $0.989$ & $0.277$ \\
\midrule
\multirow{4}{*}{PNOP-m-HO}
& Posterior-mean & $0.0700 \spm 0.0001$ & $-16630.0$ & $0.997$ & $0.024$ \\
& Prior-mean     & $0.1855 \spm 0.0936$ & $-14882.7$ & $0.936$ & $0.031$ \\
& Prior-sample   & $0.2327 \spm 0.1310$ & $-16445.0$ & $0.989$ & $0.047$ \\
& Zero-latent    & $1.9424 \spm 0.3108$ & $-3564.4$  & $0.990$ & $0.591$ \\
\midrule
\multirow{4}{*}{PNOP-HE}
& Posterior-mean & $0.0727 \spm 0.0023$ & $-16601.2$ & $0.996$ & $0.024$ \\
& Prior-mean     & $0.0730 \spm 0.0020$ & $-16598.6$ & $0.996$ & $0.024$ \\
& Prior-sample   & $0.0732 \spm 0.0018$ & $-16606.2$ & $0.996$ & $0.024$ \\
& Zero-latent    & $0.1268 \spm 0.0342$ & $-15764.2$ & $0.949$ & $0.024$ \\
\midrule
\multirow{4}{*}{PNOP-HO}
& Posterior-mean & $0.0704 \spm 0.0033$ & $-16624.5$ & $0.996$ & $0.024$ \\
& Prior-mean     & $0.0703 \spm 0.0033$ & $-16624.4$ & $0.997$ & $0.024$ \\
& Prior-sample   & $0.0705 \spm 0.0033$ & $-16627.3$ & $0.997$ & $0.024$ \\
& Zero-latent    & $0.1077 \spm 0.0298$ & $-16086.9$ & $0.968$ & $0.024$ \\
\bottomrule
\end{tabular}
}
\end{table*}

We examine the effect of how the global latent variable is injected into the probabilistic operator decoder. To this end, we compare PNOP, where the latent is injected once at the input stage before operator propagation, to a model variant, termed PNOP-m, in which the latent is injected through decoder-wide FiLM modulation, so that it modulates each FNO block via feature-wise affine transformations. This ablation isolates the latent-to-decoder interface while keeping the set encoder, the global latent prior/posterior mechanism, and the Gaussian output head otherwise comparable.

Table~\ref{tab:darcy_latent_injection_ablation} reports this comparison on Darcy, where the effect of the injection mechanism is clearer. The two variants are relatively close under posterior-mean evaluation, with PNOP-m slightly better in both variance settings. However, this advantage does not transfer to prior-conditioned inference. For PNOP-m the transition from posterior-mean to prior-mean leads to a substantial degradation in error and likelihood, with a further loss under prior sampling, especially in the HO case. This posterior test confirms that the decoder retains sufficient capacity to reconstruct the solution; the primary bottleneck is the prior inference. By contrast, PNOP remains highly stable across posterior-mean, prior-mean, and prior sample, with only negligible changes in all reported metrics. The posterior--prior gap is therefore much larger for the injection style in PNOP-m than for PNOP.

Similarly, removing the latent causes a severe degradation for PNOP-m, together with a marked increase in interval width, whereas PNOP deteriorates much less sharply. Indeed, the FiLM-based variant relies on a latent interaction that is less robust when prior conditioning is used directly or when latent information is weakened.

Therefore, decoder-wide FiLM modulation can over-couple the global latent with iterative operator propagation in demanding regimes such as Darcy. FiLM-based injection may allow the model to \emph{over-condition} on the posterior during training, leading to a failure in generalizing to the prior space at test time. In contrast, input-stage injection appears to provide a cleaner interface between global stochastic conditioning and the operator decoder, yielding substantially more stable inference across evaluation modes.

\subsection{Supplementary Comparisons with Branch--Trunk Operator Parameterizations}
\label{sec:deeponet_baselines}

As a supplementary dense-input reference for architectural context, we additionally evaluate three instantiations of the classical DeepONet branch--trunk parameterization~\citep{lu2021deeponet}. This comparison is not used to establish the main claims, but rather to place the proposed framework relative to a canonical non-spectral neural-operator family based on branch--trunk composition.

Since this comparison is included only as a supplementary dense-input reference, we keep the variants close to a standard DeepONet formulation while probing a small number of modest extensions. DeepONet-Flat follows the classical dense-input formulation, where the discretized input field is flattened and processed by a standard MLP. DeepONet-Attention uses learned query tokens that attend to the input point--value pairs through multi-head cross-attention~\citep{vaswani2017attention} without changing the classical branch--trunk factorization. DeepONet-FiLM conditions the trunk through feature-wise linear modulation generated from the branch representation, allowing the basis functions to adapt while preserving the overall branch--trunk decomposition.

Table~\ref{tab:det_uniform_with_deeponet} shows a clear separation between the sparse conditional \textsc{NOP} models and the additional DeepONet references. Across all three PDE benchmarks, \textsc{DNOP} and \textsc{DANOP} remain substantially more accurate despite operating only on sparse observations under the standard uniform-context setting. The relative ordering within the DeepONet family varies across benchmarks, but all three variants remain clearly weaker. Indeed, in the present setting, the classical branch--trunk parameterization remains less effective than the \textsc{NOP} models operating under sparse context.

\begin{table*}[t]
\caption{Deterministic rel-$L^2$ error across PDE benchmarks. Mean values over 4 seeds. \textsc{NOP} models use sparse observations under the standard uniform-context setting, whereas DeepONet variants use the full dense input.}
\label{tab:det_uniform_with_deeponet}
\centering
\setlength{\tabcolsep}{2pt}
\resizebox{0.88\linewidth}{!}{%
\begin{tabular}{l cc ccc}
\toprule
& \multicolumn{2}{c}{\textbf{Uniform Sparse Context}} & \multicolumn{3}{c}{\textbf{Dense Grid References}} \\
\cmidrule(lr){2-3} \cmidrule(lr){4-6}
\textbf{PDE} & \textbf{DNOP} & \textbf{DANOP} & \textbf{DeepONet-Flat} & \textbf{DeepONet-Attn} & \textbf{DeepONet-FiLM} \\
\midrule
Burgers 
& $0.0058 \spm 0.0001$ 
& $0.0047 \spm 0.0001$ 
& $0.1959 \spm 0.0015$ 
& $0.3339 \spm 0.0363$
& $0.1025 \spm 0.0085$ \\

Darcy   
& $0.0242 \spm 0.0002$ 
& $0.0226 \spm 0.0008$ 
& $0.1275 \spm 0.0045$ 
& $0.1165 \spm 0.0053$
& $0.0896 \spm 0.0017$ \\

NS      
& $0.0323 \spm 0.0030$ 
& $0.0326 \spm 0.0016$ 
& $0.1364 \spm 0.0011$ 
& $0.2087 \spm 0.0437$
& $0.1121 \spm 0.0081$ \\
\bottomrule
\end{tabular}%
}
\end{table*}

\section{Extended architectural and generalization ablations}
\label{app:arch_diagnostics}

\subsection{Context Strategy and Context Budget Ablations}
\label{sec:appendix_context_ablations}

% =========================================================
% COMBINED TABLE 1: DETERMINISTIC ABLATION (Tables 11 & 35)
% =========================================================

\begin{table}[htbp]
\centering
\caption{Deterministic context strategy and context budget ablation. Mean rel-$L^2$ error over 4 seeds, specific for this ablation suite.}
\label{tab:deterministic_context_ablation_combined}
\resizebox{0.83\linewidth}{!}{%
\begin{tabular}{lllccc}
\toprule
\textbf{PDE} & \textbf{Budget} & \textbf{Model} & \textbf{Uniform} & \textbf{Clustered} & \textbf{Mixed} \\
\midrule
\multirow{6}{*}{Burgers} 
    & \multirow{2}{*}{Low}  & DNOP  & $0.0054 \spm 0.0004$ & $0.0106 \spm 0.0006$ & $0.0067 \spm 0.0002$ \\
    &                       & DANOP & $0.0047 \spm 0.0003$ & $0.0057 \spm 0.0002$ & $0.0047 \spm 0.0002$ \\
\cmidrule(lr){2-6}
    & \multirow{2}{*}{Base} & DNOP  & $0.0060 \spm 0.0003$ & $0.0115 \spm 0.0003$ & $0.0067 \spm 0.0004$ \\
    &                       & DANOP & $0.0048 \spm 0.0002$ & $0.0059 \spm 0.0003$ & $0.0049 \spm 0.0001$ \\
\cmidrule(lr){2-6}
    & \multirow{2}{*}{High} & DNOP  & $0.0061 \spm 0.0005$ & $0.0118 \spm 0.0002$ & $0.0067 \spm 0.0002$ \\
    &                       & DANOP & $0.0047 \spm 0.0002$ & $0.0061 \spm 0.0002$ & $0.0048 \spm 0.0002$ \\
\midrule
\multirow{6}{*}{Darcy} 
    & \multirow{2}{*}{Low}  & DNOP  & $0.0242 \spm 0.0009$ & $0.0217 \spm 0.0005$ & $0.0199 \spm 0.0008$ \\
    &                       & DANOP & $0.0231 \spm 0.0006$ & $0.0233 \spm 0.0011$ & $0.0234 \spm 0.0007$ \\
\cmidrule(lr){2-6}
    & \multirow{2}{*}{Base} & DNOP  & $0.0236 \spm 0.0003$ & $0.0176 \spm 0.0013$ & $0.0170 \spm 0.0010$ \\
    &                       & DANOP & $0.0222 \spm 0.0009$ & $0.0232 \spm 0.0008$ & $0.0230 \spm 0.0004$ \\
\cmidrule(lr){2-6}
    & \multirow{2}{*}{High} & DNOP  & $0.0215 \spm 0.0004$ & $0.0159 \spm 0.0003$ & $0.0169 \spm 0.0006$ \\
    &                       & DANOP & $0.0227 \spm 0.0005$ & $0.0230 \spm 0.0006$ & $0.0228 \spm 0.0009$ \\
\midrule
\multirow{6}{*}{NS} 
    & \multirow{2}{*}{Low}  & DNOP  & $0.0324 \spm 0.0006$ & $0.0347 \spm 0.0007$ & $0.0353 \spm 0.0004$ \\
    &                       & DANOP & $0.0340 \spm 0.0004$ & $0.0336 \spm 0.0009$ & $0.0335 \spm 0.0008$ \\
\cmidrule(lr){2-6}
    & \multirow{2}{*}{Base} & DNOP  & $0.0293 \spm 0.0006$ & $0.0354 \spm 0.0002$ & $0.0343 \spm 0.0004$ \\
    &                       & DANOP & $0.0322 \spm 0.0003$ & $0.0336 \spm 0.0007$ & $0.0337 \spm 0.0003$ \\
\cmidrule(lr){2-6}
    & \multirow{2}{*}{High} & DNOP  & $0.0252 \spm 0.0006$ & $0.0335 \spm 0.0006$ & $0.0332 \spm 0.0002$ \\
    &                       & DANOP & $0.0295 \spm 0.0008$ & $0.0339 \spm 0.0008$ & $0.0327 \spm 0.0007$ \\
\bottomrule
\end{tabular}
}
\end{table}

% =========================================================
% COMBINED TABLE 2: PROBABILISTIC ABLATION (Tables 12, 36 & 37)
% =========================================================

\begin{table}[htbp]
\centering
\caption{Probabilistic context strategy ablation under prior-sample evaluation at the base context budget. Mean values specific for this ablation suite over 4 seeds.}
\label{tab:probabilistic_context_ablation_combined}
\resizebox{0.88\linewidth}{!}{%
\begin{tabular}{lllcccc}
\toprule
\textbf{PDE} & \textbf{Strategy} & \textbf{Model} & \textbf{Rel-$L^2$} & \textbf{MC-NLL} & \textbf{Coverage} & \textbf{Width} \\
\midrule
\multirow{12}{*}{Burgers} 
& \multirow{4}{*}{Uniform}   
& PNOP-HE  & $0.0145 \spm 0.0087$ & $-1001.6$ & $0.9745$ & $0.0472$ \\
&& PNOP-HO  & $0.0113 \spm 0.0004$ & $-859.4$ & $0.9869$ & $0.0451$ \\
&& PANOP-HE & $0.0092 \spm 0.0005$ & $-995.2$ & $0.9720$ & $0.0247$ \\
&& PANOP-HO & $0.0112 \spm 0.0001$ & $-857.1$ & $0.9873$ & $0.0460$ \\
\cmidrule{2-7}
& \multirow{4}{*}{Clustered} 
& PNOP-HE  & $0.0098 \spm 0.0005$ & $-994.7$ & $0.9695$ & $0.0248$ \\
&& PNOP-HO  & $0.0153 \spm 0.0001$ & $-766.3$ & $0.9907$ & $0.0674$ \\
&& PANOP-HE & $0.0089 \spm 0.0003$ & $-998.0$ & $0.9746$ & $0.0248$ \\
&& PANOP-HO & $0.0167 \spm 0.0006$ & $-739.3$ & $0.9921$ & $0.0755$ \\
\cmidrule{2-7}
& \multirow{4}{*}{Mixed}     
& PNOP-HE  & $0.0092 \spm 0.0001$ & $-999.2$ & $0.9730$ & $0.0247$ \\
&& PNOP-HO  & $0.0134 \spm 0.0001$ & $-809.5$ & $0.9889$ & $0.0561$ \\
&& PANOP-HE & $0.0086 \spm 0.0003$ & $-999.3$ & $0.9736$ & $0.0245$ \\
&& PANOP-HO & $0.0145 \spm 0.0002$ & $-777.0$ & $0.9917$ & $0.0650$ \\
\midrule
\multirow{12}{*}{Darcy} 
& \multirow{4}{*}{Uniform}   
& PNOP-HE  & $0.0754 \spm 0.0029$ & $-16579.7$ & $0.9948$ & $0.0239$ \\
&& PNOP-HO  & $0.0660 \spm 0.0003$ & $-16669.4$ & $0.9978$ & $0.0238$ \\
&& PANOP-HE & $0.0754 \spm 0.0072$ & $-16621.9$ & $0.9964$ & $0.0252$ \\
&& PANOP-HO & $0.0656 \spm 0.0001$ & $-16672.1$ & $0.9978$ & $0.0238$ \\
\cmidrule{2-7}
& \multirow{4}{*}{Clustered} 
& PNOP-HE  & $0.0711 \spm 0.0045$ & $-16654.7$ & $0.9971$ & $0.0242$ \\
&& PNOP-HO  & $0.0640 \spm 0.0005$ & $-16687.2$ & $0.9982$ & $0.0238$ \\
&& PANOP-HE & $0.0674 \spm 0.0002$ & $-16656.4$ & $0.9976$ & $0.0238$ \\
&& PANOP-HO & $0.0654 \spm 0.0012$ & $-16673.6$ & $0.9979$ & $0.0238$ \\
\cmidrule{2-7}
& \multirow{4}{*}{Mixed}     
& PNOP-HE  & $0.0696 \spm 0.0011$ & $-16641.0$ & $0.9969$ & $0.0239$ \\
&& PNOP-HO  & $0.0641 \spm 0.0003$ & $-16686.4$ & $0.9982$ & $0.0238$ \\
&& PANOP-HE & $0.0671 \spm 0.0011$ & $-16660.5$ & $0.9976$ & $0.0239$ \\
&& PANOP-HO & $0.0655 \spm 0.0002$ & $-16672.8$ & $0.9979$ & $0.0238$ \\
\midrule
\multirow{12}{*}{NS} 
& \multirow{4}{*}{Uniform}   
& PNOP-HE  & $0.0548 \spm 0.0028$ & $-12694.2$ & $0.9010$ & $0.0634$ \\
&& PNOP-HO  & $0.1271 \spm 0.0122$ & $-8155.2$ & $0.9094$ & $0.1196$ \\
&& PANOP-HE & $0.0478 \spm 0.0016$ & $-12818.9$ & $0.9016$ & $0.0573$ \\
&& PANOP-HO & $0.1128 \spm 0.0078$ & $-8311.0$ & $0.8944$ & $0.1026$ \\
\cmidrule{2-7}
& \multirow{4}{*}{Clustered} 
& PNOP-HE  & $0.0729 \spm 0.0032$ & $-11812.0$ & $0.9041$ & $0.0867$ \\
&& PNOP-HO  & $0.2038 \spm 0.0144$ & $-6509.3$ & $0.9193$ & $0.1997$ \\
&& PANOP-HE & $0.0563 \spm 0.0020$ & $-12332.2$ & $0.9056$ & $0.0695$ \\
&& PANOP-HO & $0.1735 \spm 0.0304$ & $-6917.0$ & $0.9165$ & $0.1676$ \\
\cmidrule{2-7}
& \multirow{4}{*}{Mixed}     
& PNOP-HE  & $0.0751 \spm 0.0141$ & $-11676.2$ & $0.8833$ & $0.0782$ \\
&& PNOP-HO  & $0.2186 \spm 0.0216$ & $-6470.3$ & $0.9106$ & $0.1909$ \\
&& PANOP-HE & $0.0763 \spm 0.0319$ & $-12377.7$ & $0.9180$ & $0.0955$ \\
&& PANOP-HO & $0.1872 \spm 0.0104$ & $-6878.6$ & $0.9240$ & $0.1831$ \\
\bottomrule
\end{tabular}
}
\end{table}

To isolate the effects of architectural choices from the intrinsic properties of the observation distributions, we evaluate the NOP family across varying context sampling strategies and additional context budgets. This allows us to characterize how conditional operator learning under sparse partial observations depends on both the geometry and the amount of available context.

\paragraph{Deterministic Analysis and the Stabilizing Role of Attention.} 
Table \ref{tab:deterministic_context_ablation_combined} summarizes the deterministic performance across benchmarks. We observe a geometric sensitivity that varies with the underlying physical problem. In Burgers 1D, the models exhibit a weak dependence on the context budget and consistently favor uniform coverage. The dynamics in Navier--Stokes reflect a similar preference for uniform distributions; increasing the context budget under non-uniform strategies (clustered or mixed) yields only limited benefits, suggesting that information propagation in this periodic domain relies more on broad spatial coverage than on local observation density. This is consistent with the interpretation that when the underlying PDE is smooth, periodic, and spectrally aligned with the FNO decoder, global observation coverage is sufficient to construct an accurate continuous representation.

Conversely, Darcy 2D exposes a much stronger dependence on observation topology. The deterministic baseline, \textsc{DNOP}, degrades under uniform sampling but improves meaningfully under clustered or mixed observations, with the sampling strategy remaining the more decisive factor even as context budget increases. Increasing the context budget does help \textsc{DNOP} within each strategy, particularly in the more favorable clustered and mixed settings, but it does not eliminate the performance gap induced by the sampling pattern itself. Thus, both context amount and context geometry matter in Darcy, with the sampling strategy remaining the more decisive factor.

The introduction of the query-aligned attention pathway in \textsc{DANOP} noticeably flattens the performance variation across strategies and budgets. In Darcy, \textsc{DANOP} mitigates much of the fragility observed in \textsc{DNOP} under suboptimal sampling distributions. Importantly, \textsc{DANOP} does not surpass the absolute best result of \textsc{DNOP}, achieved under the favorable clustered, high-budget regime, but instead yields a more robust architecture whose performance varies much less with the spatial distribution of observations. This suggests that explicitly preserving local geometry reduces the model's reliance on favorable sampling configurations.

\paragraph{Probabilistic Analysis and Variance Regularization.} 
As summarized in Table \ref{tab:probabilistic_context_ablation_combined}, on Burgers the \textsc{PANOP} variant with a HE head consolidates itself as the most accurate architecture. For Navier--Stokes, it similarly demonstrates strong performance under uniform context, yielding the best accuracy and the sharpest uncertainty estimates; across our experiments, this variant also showed the strongest prior--posterior alignment among the probabilistic models.

The Darcy scenario, however, reveals more nuanced behavior regarding the parameterization of predictive uncertainty. The HE variants maintain a residual sensitivity to the context strategy, continuing to benefit modestly from non-uniform distributions. This is consistent with previous results suggesting that the additional flexibility of spatially varying predictive variance can partially absorb observation-geometry effects rather than requiring all improvements to come from the predictive mean alone. By contrast, restricting the models to a HO predictive variance largely removes the dependence on sampling strategy. As also observed in previous sections, although a HO head sacrifices the localized adaptivity of HE uncertainty, it can act as an implicit regularizer, encouraging the probabilistic model to rely more consistently on the representational capacity of its geometry-preserving deterministic trunk (\textsc{PANOP}). In turn, this yields a stable predictive performance across uniform, clustered, and mixed strategies.

\subsection{Test-Time Context Scaling}
\label{sec:test_time_scaling}

% =========================================================
% COMBINED TABLE 1: DETERMINISTIC TEST-TIME SCALING (Tables 14 & 46)
% =========================================================
\begin{table}[htbp]
\centering
\caption{Deterministic test-time context scaling across benchmarks. Mean values over 4 seeds. Models are evaluated by changing only the inference-time context budget while reusing checkpoints trained under the base context regime. Values reported represent rel-$L^2$ error and are specific to the fixed-checkpoint scaling evaluation in this subsection.}
\label{tab:test_time_scaling_det_combined}
\resizebox{0.8\linewidth}{!}{%
\begin{tabular}{llccc}
\toprule
\textbf{PDE} & \textbf{Model} & \textbf{Low Test Budget} & \textbf{Base Test Budget} & \textbf{High Test Budget} \\
\midrule
\multirow{2}{*}{Burgers} 
& DNOP & $0.00722 \spm 0.00053$ & $0.00598 \spm 0.00033$ & $0.01042 \spm 0.00082$ \\
& DANOP & $0.00544 \spm 0.00022$ & $0.00476 \spm 0.00017$ & $0.00481 \spm 0.00019$ \\
\midrule
\multirow{2}{*}{Darcy} 
& DNOP & $0.02391 \spm 0.00032$ & $0.02354 \spm 0.00032$ & $0.02474 \spm 0.00047$ \\
& DANOP & $0.02229 \spm 0.00086$ & $0.02222 \spm 0.00088$ & $0.02222 \spm 0.00086$ \\
\midrule
\multirow{2}{*}{NS} 
& DNOP & $0.03097 \spm 0.00088$ & $0.02925 \spm 0.00055$ & $0.03412 \spm 0.00030$ \\
& DANOP & $0.03265 \spm 0.00048$ & $0.03224 \spm 0.00033$ & $0.03230 \spm 0.00039$ \\
\bottomrule
\end{tabular}
}
\end{table}

% =========================================================
% COMBINED TABLE 2: PROBABILISTIC TEST-TIME SCALING (Tables 15, 16, 47 & 48)
% =========================================================
\begin{table}[htbp]
\centering
\caption{Probabilistic test-time context scaling under prior-sample evaluation. All models were trained at the base context budget using a uniform context strategy. Mean values over 4 seeds.}
\label{tab:test_time_scaling_prob_combined}
\resizebox{0.88\linewidth}{!}{%
\begin{tabular}{lllcccc}
\toprule
\textbf{PDE} & \textbf{Test Budget} & \textbf{Model} & \textbf{Rel-$L^2$} & \textbf{MC-NLL} & \textbf{Coverage} & \textbf{Width} \\
\midrule
\multirow{12}{*}{Burgers}
& \multirow{4}{*}{Low}  & PNOP-HE  & $0.0311 \pm 0.0087$ & $-966.5$  & $0.9651$ & $0.0421$ \\
&                        & PNOP-HO  & $0.0119 \pm 0.0004$ & $-860.8$  & $0.9883$ & $0.0463$ \\
&                        & PANOP-HE & $0.0092 \pm 0.0005$ & $-995.4$  & $0.9716$ & $0.0250$ \\
&                        & PANOP-HO & $0.0114 \pm 0.0001$ & $-861.9$  & $0.9873$ & $0.0463$ \\
\cmidrule{2-7}
& \multirow{4}{*}{Base} & PNOP-HE  & $0.0165 \pm 0.0005$ & $-988.5$  & $0.9689$ & $0.0274$ \\
&                        & PNOP-HO  & $0.0113 \pm 0.0004$ & $-863.9$  & $0.9873$ & $0.0462$ \\
&                        & PANOP-HE & $0.0092 \pm 0.0005$ & $-995.2$  & $0.9722$ & $0.0247$ \\
&                        & PANOP-HO & $0.0113 \pm 0.0001$ & $-862.1$  & $0.9870$ & $0.0462$ \\
\cmidrule{2-7}
& \multirow{4}{*}{High} & PNOP-HE  & $0.0205 \pm 0.0005$ & $-977.7$  & $0.9669$ & $0.0311$ \\
&                        & PNOP-HO  & $0.0139 \pm 0.0004$ & $-855.3$  & $0.9876$ & $0.0461$ \\
&                        & PANOP-HE & $0.0100 \pm 0.0005$ & $-986.7$  & $0.9663$ & $0.0247$ \\
&                        & PANOP-HO & $0.0117 \pm 0.0001$ & $-860.4$  & $0.9872$ & $0.0461$ \\
\midrule
\multirow{12}{*}{Darcy}
& \multirow{4}{*}{Low}  & PNOP-HE  & $0.0814 \pm 0.0029$ & $-16514.0$ & $0.9950$ & $0.0238$ \\
&                        & PNOP-HO  & $0.0660 \pm 0.0003$ & $-16668.8$ & $0.9980$ & $0.0238$ \\
&                        & PANOP-HE & $0.0802 \pm 0.0036$ & $-16528.1$ & $0.9948$ & $0.0238$ \\
&                        & PANOP-HO & $0.0657 \pm 0.0001$ & $-16671.5$ & $0.9980$ & $0.0239$ \\
\cmidrule{2-7}
& \multirow{4}{*}{Base} & PNOP-HE  & $0.0755 \pm 0.0029$ & $-16575.7$ & $0.9960$ & $0.0238$ \\
&                        & PNOP-HO  & $0.0658 \pm 0.0003$ & $-16670.0$ & $0.9979$ & $0.0238$ \\
&                        & PANOP-HE & $0.0721 \pm 0.0036$ & $-16606.8$ & $0.9965$ & $0.0238$ \\
&                        & PANOP-HO & $0.0657 \pm 0.0001$ & $-16671.8$ & $0.9979$ & $0.0239$ \\
\cmidrule{2-7}
& \multirow{4}{*}{High} & PNOP-HE  & $0.0770 \pm 0.0029$ & $-16560.6$ & $0.9963$ & $0.0238$ \\
&                        & PNOP-HO  & $0.0661 \pm 0.0003$ & $-16667.1$ & $0.9979$ & $0.0238$ \\
&                        & PANOP-HE & $0.0716 \pm 0.0036$ & $-16611.8$ & $0.9963$ & $0.0238$ \\
&                        & PANOP-HO & $0.0657 \pm 0.0001$ & $-16671.4$ & $0.9979$ & $0.0239$ \\
\midrule
\multirow{12}{*}{NS}
& \multirow{4}{*}{Low}  & PNOP-HE  & $0.0580 \pm 0.0028$ & $-11986.3$ & $0.8936$ & $0.0644$ \\
&                        & PNOP-HO  & $0.1336 \pm 0.0122$ & $-7072.5$  & $0.9067$ & $0.1199$ \\
&                        & PANOP-HE & $0.0488 \pm 0.0016$ & $-12632.5$ & $0.9046$ & $0.0572$ \\
&                        & PANOP-HO & $0.1117 \pm 0.0078$ & $-7310.1$  & $0.8960$ & $0.1025$ \\
\cmidrule{2-7}
& \multirow{4}{*}{Base} & PNOP-HE  & $0.0548 \pm 0.0028$ & $-12131.0$ & $0.9010$ & $0.0634$ \\
&                        & PNOP-HO  & $0.1271 \pm 0.0122$ & $-7128.9$  & $0.9094$ & $0.1196$ \\
&                        & PANOP-HE & $0.0478 \pm 0.0016$ & $-12608.4$ & $0.9016$ & $0.0573$ \\
&                        & PANOP-HO & $0.1128 \pm 0.0078$ & $-7368.9$  & $0.8944$ & $0.1026$ \\
\cmidrule{2-7}
& \multirow{4}{*}{High} & PNOP-HE  & $0.0681 \pm 0.0028$ & $-11616.1$ & $0.8984$ & $0.0637$ \\
&                        & PNOP-HO  & $0.1769 \pm 0.0122$ & $-6527.9$  & $0.9048$ & $0.1196$ \\
&                        & PANOP-HE & $0.0495 \pm 0.0022$ & $-12513.9$ & $0.9022$ & $0.0579$ \\
&                        & PANOP-HO & $0.1116 \pm 0.0078$ & $-7345.7$  & $0.8948$ & $0.1029$ \\
\bottomrule
\end{tabular}
}
\end{table}

The previous ablations showed that the context distribution during training significantly impacts the learned representation. A natural follow-up question is whether simply providing more context observations at inference time, while reusing a fixed checkpoint trained under a specific context regime, yields systematic performance gains. To isolate this effect, we evaluate models trained at the base context budget across three test-time budgets to assess inference-time scaling behavior across varied physical dynamics.

\paragraph{Deterministic Behavior and Checkpoint Specialization.} 
Table \ref{tab:test_time_scaling_det_combined} summarizes the test-time context scaling results for the deterministic architectures. The most salient observation across all three benchmarks is that increasing the context budget exclusively at test time does not yield monotonic performance improvements. For \textsc{DNOP}, the base budget used during training is typically best, or statistically indistinguishable from the best setting, while moving to the high test-time budget either leaves performance unchanged or degrades it, most visibly in Burgers and Navier--Stokes. This suggests that the baseline checkpoint remains adapted to the observation density seen during training and does not systematically benefit from denser test-time conditioning alone.

In contrast, \textsc{DANOP} architecture demonstrates substantially greater stability. Across Burgers, Darcy, and Navier--Stokes, the relative-$L^2$ error for \textsc{DANOP} remains nearly unchanged across the low, base, and high evaluation budgets. In some cases, the high-budget regime matches the base budget or yields only a slight improvement, but these effects are small and not systematic. This suggests that once a robust conditional representation has been learned, additional test-time context can be assimilated without destabilizing inference, yet does not reliably translate into further gains in predictive accuracy.

\paragraph{Probabilistic Behavior and Uncertainty Stability.} 
Table \ref{tab:test_time_scaling_prob_combined} extends this analysis to the probabilistic models evaluated under prior-sampled inference. The results broadly reinforce the deterministic findings while highlighting differences in architectural robustness. The strongest probabilistic variants, particularly \textsc{PANOP}, remain stable across test-time budgets in most settings. In Burgers and Navier--Stokes, \textsc{PANOP}-HE shows only minor variation in predictive accuracy, empirical coverage, and average interval width as the inference budget changes.

A more nuanced pattern appears in Darcy. The HO variants of both \textsc{PNOP} and \textsc{PANOP} are almost completely invariant to the test-time context budget, with negligible differences across all reported metrics. By contrast, the HE models retain some sensitivity, and \textsc{PANOP}-HE shows a visible improvement when moving from the low-budget regime to the base or high-budget regimes. Thus, \textsc{PANOP} remains generally stable overall, with near invariance in the HO setting, while Darcy with HE variance still benefits modestly from moving away from the low-budget regime.

Conversely, the global-latent \textsc{PNOP} architecture with a HO head exhibits greater fragility. In Navier--Stokes, increasing the test-time context to the high-budget regime causes a clear degradation in predictive accuracy without correspondingly sharpening predictive intervals. This suggests that weaker probabilistic designs do not automatically become robust simply by receiving more observations at inference time.

Overall, these results suggest that performance is governed primarily by the conditional representation learned during training, rather than improving automatically with additional test-time context at deployment. Architectures that preserve local geometry effectively, such as \textsc{DANOP} and \textsc{PANOP}, can absorb moderate test-time changes in context budget without substantial degradation, but they do not exhibit systematic gains from extra context alone.

\subsection{Train-Evaluation Strategy}
\label{sec:train_eval_matrix}

% =========================================================
% COMBINED TABLE 1: DETERMINISTIC TRAIN-EVAL MATRIX 
% (Tables 20, 23, & 38)
% =========================================================

\begin{table}[htbp]
\centering
\caption{Deterministic train--evaluation context strategy matrix across benchmarks. Each entry reports mean rel-$L^2$ error over 4 seeds evaluated at the base context budget for this ablation.}
\label{tab:train_eval_matrix_det_combined}
\resizebox{0.9\linewidth}{!}{%
\begin{tabular}{lllccc}
\toprule
\textbf{PDE} & \textbf{Model} & \textbf{Train Strategy} & \textbf{Eval: Uniform} & \textbf{Eval: Mixed} & \textbf{Eval: Clustered} \\
\midrule
\multirow{6}{*}{Burgers}
& \multirow{3}{*}{DNOP}
& Uniform   & $0.0060 \pm 0.0003$ & $0.0119 \pm 0.0008$ & $0.0175 \pm 0.0008$ \\
&           & Mixed     & $0.0069 \pm 0.0002$ & $0.0069 \pm 0.0002$ & $0.0068 \pm 0.0004$ \\
&           & Clustered & $0.0140 \pm 0.0008$ & $0.0104 \pm 0.0008$ & $0.0073 \pm 0.0004$ \\
\cmidrule(lr){2-6}
& \multirow{3}{*}{DANOP}
& Uniform   & $0.0048 \pm 0.0002$ & $0.0066 \pm 0.0002$ & $0.0092 \pm 0.0002$ \\
&           & Mixed     & $0.0049 \pm 0.0001$ & $0.0049 \pm 0.0001$ & $0.0049 \pm 0.0001$ \\
&           & Clustered & $0.0050 \pm 0.0004$ & $0.0050 \pm 0.0004$ & $0.0050 \pm 0.0004$ \\
\midrule
\multirow{6}{*}{Darcy}
& \multirow{3}{*}{DNOP}
& Uniform   & $0.0235 \pm 0.0003$ & $0.0436 \pm 0.0005$ & $0.0582 \pm 0.0005$ \\
&           & Mixed     & $0.0170 \pm 0.0010$ & $0.0170 \pm 0.0010$ & $0.0171 \pm 0.0010$ \\
&           & Clustered & $0.0178 \pm 0.0013$ & $0.0177 \pm 0.0013$ & $0.0176 \pm 0.0013$ \\
\cmidrule(lr){2-6}
& \multirow{3}{*}{DANOP}
& Uniform   & $0.0222 \pm 0.0009$ & $0.0236 \pm 0.0005$ & $0.0244 \pm 0.0009$ \\
&           & Mixed     & $0.0230 \pm 0.0004$ & $0.0230 \pm 0.0004$ & $0.0230 \pm 0.0004$ \\
&           & Clustered & $0.0232 \pm 0.0008$ & $0.0232 \pm 0.0008$ & $0.0232 \pm 0.0008$ \\
\midrule
\multirow{6}{*}{NS}
& \multirow{3}{*}{DNOP}
& Uniform   & $0.0292 \pm 0.0005$ & $0.0918 \pm 0.0002$ & $0.1337 \pm 0.0002$ \\
&           & Mixed     & $0.0340 \pm 0.0009$ & $0.0343 \pm 0.0004$ & $0.0343 \pm 0.0004$ \\
&           & Clustered & $0.0366 \pm 0.0002$ & $0.0360 \pm 0.0002$ & $0.0354 \pm 0.0002$ \\
\cmidrule(lr){2-6}
& \multirow{3}{*}{DANOP}
& Uniform   & $0.0322 \pm 0.0003$ & $0.0407 \pm 0.0004$ & $0.0466 \pm 0.0004$ \\
&           & Mixed     & $0.0336 \pm 0.0003$ & $0.0337 \pm 0.0003$ & $0.0337 \pm 0.0003$ \\
&           & Clustered & $0.0335 \pm 0.0007$ & $0.0335 \pm 0.0007$ & $0.0335 \pm 0.0007$ \\
\bottomrule
\end{tabular}
}
\end{table}

\begin{figure}[t]
\centering
\includegraphics[width=\linewidth]{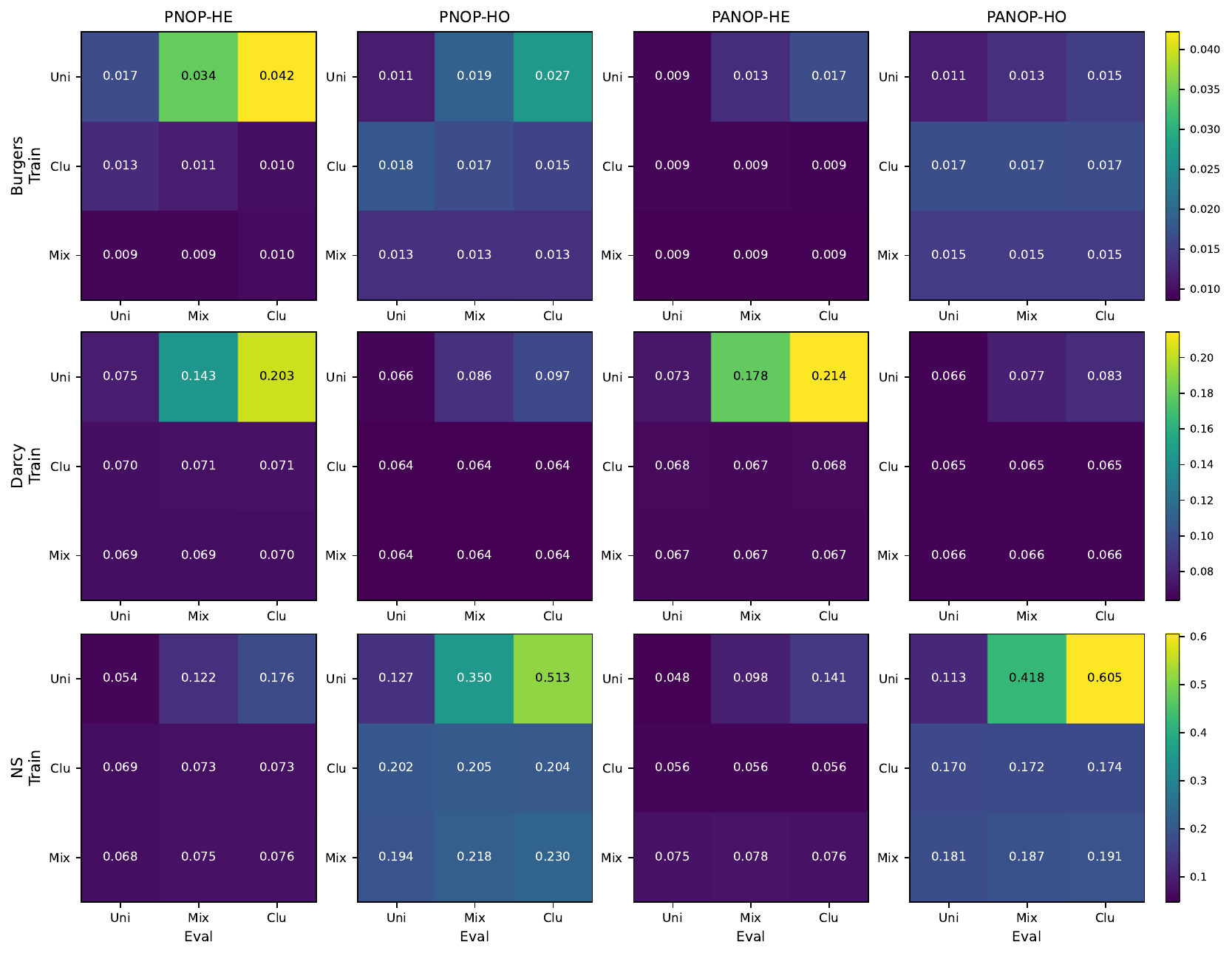}
\caption{\textbf{Probabilistic train--evaluation rel-$L^2$ matrix under prior-sample evaluation.} Each cell shows performance for a train-eval strategy pair, with rows indicating training distribution and columns evaluation distribution. The diagonal corresponds to matched train/eval geometry. PNOP shows pronounced off-diagonal degradation when trained on uniform contexts, particularly in Darcy and NS, whereas PANOP (especially HE) substantially flattens the matrix, indicating stronger robustness to sensor-placement shift. Per-PDE normalization is used to make patterns visible across regimes of different scale.}
\label{fig:train_eval_heatmap_prob}
\vspace{-0.5em}
\end{figure}

To determine whether the models learn a robust conditional representation or specialize to the context geometry observed during training, we evaluate their performance under out-of-distribution (OOD) geometry shifts. For each benchmark, models are trained under three context sampling strategies and each resulting checkpoint is evaluated across all three strategies at a fixed base budget. 

\paragraph{Deterministic Robustness and Specialization.} 
As shown in Table \ref{tab:train_eval_matrix_det_combined}, the baseline \textsc{DNOP} exhibits train--eval specialization across all three physical domains. When trained on uniform contexts, \textsc{DNOP} achieves strong in-distribution results but degrades under geometry shift, with the largest penalties typically appearing when evaluation moves to clustered observations and already sizable degradation under mixed evaluation. This effect is most severe in Navier--Stokes and also pronounced in Darcy, indicating that the SetConv-only representation remains tightly coupled to the spatial coverage pattern seen during training.

By contrast, training with mixed or clustered contexts yields much flatter evaluation rows, indicating a trade-off between specialization and robustness in Burgers and Navier--Stokes. In Darcy, however, the pattern is even stronger showing that for \textsc{DNOP} training on mixed or clustered contexts not only improves robustness to evaluation-time geometry shifts, but also improves absolute performance relative to training on uniform contexts, including under uniform evaluation itself. Thus, in Darcy, non-uniform training geometries are not merely a robustness device, they appear better aligned with the structure of the task.

The introduction of the attention-based local trunk in \textsc{DANOP} largely reduces this tension. Across Burgers and Darcy, \textsc{DANOP} substantially flattens the evaluation matrix, effectively removing the severe performance drop under strategy mismatch. In Navier--Stokes, \textsc{DANOP} likewise reduces the severity of the cross-strategy penalty, although some residual sensitivity remains when training and evaluation geometries differ sharply. Overall, these results support the view that query-aligned geometry preservation is an effective architectural mechanism for reducing sensitivity to sensor-placement shift.

\paragraph{Probabilistic Robustness and Calibration Shift.} 
%\ref{tab:train_eval_matrix_prob_combined_final}

Figure \ref{fig:train_eval_heatmap_prob} extends this analysis to the probabilistic models, showing that train--eval mismatch affects not only the predictive mean but also uncertainty quality. For the PNOP architecture, HE variants are highly sensitive to mismatch. When trained on uniform contexts and evaluated on clustered data, \textsc{PNOP} suffers a severe degradation in rel-$L^2$ error alongside a drop in empirical coverage. 

The geometry-preserving \textsc{PANOP} mitigates part of this fragility. In Darcy, its HE variant is clearly more stable than \textsc{PNOP}, especially once training uses mixed or clustered contexts. In Navier--Stokes, however, residual sensitivity remains under the strongest train--eval mismatches, particularly when transferring from uniform training to non-uniform evaluation. For instance, PANOP-HE trained on uniform contexts and evaluated on clustered observations reaches a significant relative-$L^2$ degradation compared to performance in-distribution, highlighting that geometry mismatch in in this turbulent regime remains an open challenge even for the strongest architecture.

Constraining the predictive variance to be HO yields the most stable OOD behavior overall. In Darcy, the \textsc{PANOP}-HO is remarkably flat across all train--test pairs once training includes mixed or clustered contexts, and \textsc{PNOP}-HO shows a similar pattern. In Navier--Stokes, the same broad trade-off observed in the deterministic models persists: uniform-trained probabilistic models yield the best in-distribution accuracy and relatively sharp intervals, but remain fragile under evaluation-time geometry shifts, whereas models trained on non-uniform geometries, particularly mixed and clustered regimes, maintain steadier performance and more stable coverage across evaluation strategies.

This ablation shows that observation geometry acts as a strong inductive bias during training. Uniform sampling can be favorable when deployment-time sensing is known to remain somewhat dense and evenly distributed. Training on more diverse non-uniform geometries, particularly mixed and clustered regimes, yields substantially stronger robustness across sensor configurations. Architecturally, explicitly preserving local context--query interactions (\textsc{DANOP} and \textsc{PANOP}) is important for improving generalization across observation patterns and for reducing both deterministic error spikes and probabilistic calibration degradation under train--eval mismatch.

\subsection{Cross-Resolution Transfer}
\label{sec:cross_resolution}

% =========================================================
% COMBINED TABLE 1: DETERMINISTIC CROSS-RESOLUTION TRANSFER 
% (Combined from Tables 17 & 42)
% =========================================================
\begin{table}[htbp]
\centering
\caption{Deterministic cross-resolution transfer across benchmarks. Models trained on coarse grids are evaluated on paired instances to isolate resolution shift from dataset shift. Evaluation protocols include in-distribution (Coarse $\rightarrow$ Coarse), zero-shot super-resolution (Coarse $\rightarrow$ Fine), and super-resolution with proportionally scaled context (sc). All results refer to the paired transfer benchmark designed for this ablation. Mean rel-$L^2$ over 4 seeds.}
\label{tab:cross_res_det_combined}
\resizebox{0.83\linewidth}{!}{%
\begin{tabular}{llccc}
\toprule
\textbf{PDE} & \textbf{Model} & \textbf{Coarse $\rightarrow$ Coarse} & \textbf{Coarse $\rightarrow$ Fine} & \textbf{Coarse $\rightarrow$ Fine (sc)} \\
\midrule
\multirow{2}{*}{Burgers}
& DNOP & $0.0058 \pm 0.0001$ & $0.0057 \pm 0.0001$ & $0.0085 \pm 0.0013$ \\
& DANOP & $0.0048 \pm 0.0001$ & $0.0048 \pm 0.0001$ & $0.0048 \pm 0.0001$ \\
\midrule
\multirow{2}{*}{Darcy}
& DNOP & $0.0250 \pm 0.0006$ & $0.3316 \pm 0.0131$ & $0.3287 \pm 0.0081$ \\
& DANOP & $0.0234 \pm 0.0008$ & $0.3196 \pm 0.0167$ & $0.3193 \pm 0.0164$ \\
\midrule
\multirow{2}{*}{NS}
& DNOP & $0.0187 \pm 0.0008$ & $0.2195 \pm 0.0033$ & $0.3329 \pm 0.0291$ \\
& DANOP & $0.0191 \pm 0.0006$ & $0.2031 \pm 0.0045$ & $0.2042 \pm 0.0060$ \\
\bottomrule
\end{tabular}
}
\end{table}

\begin{table}[htbp]
\centering
\caption{Probabilistic cross-resolution transfer under prior-sample evaluation on paired datasets. Mean values over 4 seeds. Avg. MC-NLL is normalized by the number of query points for comparability across resolutions. All results refer to the paired transfer benchmark designed for this ablation.}
\label{tab:cross_res_prob_combined_final}
\resizebox{0.9\linewidth}{!}{%
\begin{tabular}{lllcccc}
\toprule
\textbf{PDE} & \textbf{Transfer Setting} & \textbf{Model} & \textbf{Rel-$L^2$} & \textbf{Avg. MC-NLL} & \textbf{Coverage} & \textbf{Width} \\
\midrule
\multirow{9}{*}{Burgers} 
& \multirow{3}{*}{Coarse $\rightarrow$ Coarse} 
& PNOP-HE  & $0.0097$ & $-3.896$ & $0.9711$ & $0.0250$ \\
&                                                & PANOP-HE & $0.0089$ & $-3.915$ & $0.9743$ & $0.0249$ \\
&                                                & PANOP-HO & $0.0110$ & $-3.366$ & $0.9879$ & $0.0454$ \\
\cmidrule{2-7}
& \multirow{3}{*}{Coarse $\rightarrow$ Fine}   
& PNOP-HE  & $0.0097$ & $-3.899$ & $0.9716$ & $0.0250$ \\
&                                                & PANOP-HE & $0.0089$ & $-3.913$ & $0.9733$ & $0.0249$ \\
&                                                & PANOP-HO & $0.0110$ & $-3.367$ & $0.9878$ & $0.0453$ \\
\cmidrule{2-7}
& \multirow{3}{*}{Coarse $\rightarrow$ Fine (sc)} 
& PNOP-HE  & $0.0126$ & $-3.711$ & $0.9504$ & $0.0250$ \\
&                                                & PANOP-HE & $0.0094$ & $-3.895$ & $0.9706$ & $0.0248$ \\
&                                                & PANOP-HO & $0.0114$ & $-3.355$ & $0.9861$ & $0.0454$ \\
\midrule
\multirow{9}{*}{Darcy} 
& \multirow{3}{*}{Coarse $\rightarrow$ Coarse} 
& PNOP-HE  & $0.0727$ & $-4.055$ & $0.9958$ & $0.0239$ \\
&                                                & PANOP-HE & $0.0707$ & $-4.059$ & $0.9963$ & $0.0238$ \\
&                                                & PANOP-HO & $0.0674$ & $-4.066$ & $0.9973$ & $0.0238$ \\
\cmidrule{2-7}
& \multirow{3}{*}{Coarse $\rightarrow$ Fine}   
& PNOP-HE  & $0.3809$ & $-1.624$ & $0.5512$ & $0.0239$ \\
&                                                & PANOP-HE & $0.3568$ & $-1.898$ & $0.5713$ & $0.0239$ \\
&                                                & PANOP-HO & $0.3403$ & $-2.069$ & $0.6025$ & $0.0238$ \\
\cmidrule{2-7}
& \multirow{3}{*}{Coarse $\rightarrow$ Fine (sc)} 
& PNOP-HE  & $0.3871$ & $-1.512$ & $0.5487$ & $0.0239$ \\
&                                                & PANOP-HE & $0.3616$ & $-1.836$ & $0.5658$ & $0.0239$ \\
&                                                & PANOP-HO & $0.3404$ & $-2.064$ & $0.6024$ & $0.0238$ \\
\midrule
\multirow{12}{*}{NS} 
& \multirow{4}{*}{Coarse $\rightarrow$ Coarse} 
& PNOP-HE  & $0.0982$ & $-3.857$ & $0.9595$ & $0.0263$ \\
&                                                & PANOP-HE & $0.0973$ & $-3.869$ & $0.9598$ & $0.0249$ \\
&                                                & PNOP-HO  & $0.4047$ & $-2.518$ & $0.9898$ & $0.1087$ \\
&                                                & PANOP-HO & $0.3885$ & $-2.546$ & $0.9910$ & $0.1067$ \\
\cmidrule{2-7}
& \multirow{4}{*}{Coarse $\rightarrow$ Fine}   
& PNOP-HE  & $0.2222$ & $-0.367$ & $0.7087$ & $0.0372$ \\
&                                                & PANOP-HE & $0.2585$ & $3.344$  & $0.6503$ & $0.0337$ \\
&                                                & PNOP-HO  & $0.4481$ & $-2.141$ & $0.9240$ & $0.1089$ \\
&                                                & PANOP-HO & $0.4291$ & $-2.048$ & $0.9197$ & $0.1067$ \\
\cmidrule{2-7}
& \multirow{4}{*}{Coarse $\rightarrow$ Fine (sc)} 
& PNOP-HE  & $0.4385$ & $3.622$  & $0.5636$ & $0.0301$ \\
&                                                & PANOP-HE & $0.3037$ & $3.636$  & $0.6188$ & $0.0323$ \\
&                                                & PNOP-HO  & $1.3305$ & $-0.885$ & $0.7152$ & $0.1092$ \\
&                                                & PANOP-HO & $0.5493$ & $-2.014$ & $0.9026$ & $0.1067$ \\
\bottomrule
\end{tabular}
}
\end{table}

A core motivation for neural operator learning is the capacity to learn discretization-invariant mappings that can transfer zero-shot to higher resolutions. However, learning operators from sparse, ungridded context observations introduces a structural challenge as the model must construct a continuous representation from discrete points before evaluating the operator. To assess whether the learned conditional representations transfer robustly across spatial discretizations without retraining, we evaluate models trained on coarse grids under three inference protocols: paired coarse (in-distribution), paired fine (zero-shot transfer), and paired fine scaled (where the number of context points is proportionally increased to maintain relative observation density). Coarse and fine grids are $256\!\to\!512$ for Burgers and $64^2\!\to\!128^2$ for Darcy and Navier--Stokes.

\paragraph{Deterministic Transfer and the Boundaries of Mesh Invariance.} 
As shown in Table \ref{tab:cross_res_det_combined}, cross-resolution behavior is highly benchmark-dependent, clarifying the empirical limits of mesh invariance in conditional operator learning. On the 1D Burgers equation, zero-shot transfer is largely successful. The baseline \textsc{DNOP} maintains low relative error across resolutions, and the geometry-preserving \textsc{DANOP} is nearly invariant, achieving a similar rel-$L^2$ error under all three evaluation settings. Thus, in a smooth periodic setting well aligned with the FNO decoder, the learned conditional representation can transfer cleanly across resolutions from sparse observations.

By contrast, the 2D non-periodic and multi-scale settings expose a limitation. In Darcy, both deterministic models degrade upon transfer, with relative error increasing from the paired coarse setting to the paired fine grid. Navier--Stokes shows a similarly strong degradation. Scaling the context budget proportionally at the fine resolution does not rescue performance in either 2D benchmark. Thus, the issue is not simply a lack of spatial point density at inference time, but that the learned conditional representation does not transfer cleanly across discretizations in these settings. For elliptic non-periodic problems and more complex fluid dynamics, achieving robust discretization transfer under sparse partial observations remains challenging.

The contrast between Burgers and the 2D benchmarks is consistent with the possibility that the FNO decoder transfers more readily in smooth, spectrally aligned settings than in non-periodic or more spatially complex regimes. However, this ablation does not isolate the decoder from the conditional representation pipeline, so the observed cross-resolution failures should be interpreted as limitations of the overall sparse conditional operator architecture rather than of the FNO backbone alone.

\paragraph{Probabilistic Transfer and Calibration Shift.} 
Results in Table \ref{tab:cross_res_prob_combined_final} reinforce the deterministic findings and highlight the vulnerability of uncertainty estimation under resolution shift. For Burgers 1D, where the deterministic mapping transfers well, the probabilistic models also generalize reliably. \textsc{PANOP} with a HE likelihood retains strong accuracy, stable MC-NLL, and near-nominal empirical coverage across all coarse and fine evaluation protocols.

However, in Darcy and Navier--Stokes, zero-shot resolution transfer induces both an accuracy failure and a marked calibration shift. For instance, the \textsc{PANOP}-HE drops severely in coverage from the coarse grid to the fine grid. The average interval widths remain nearly unchanged in Darcy and change only modestly in several Navier--Stokes settings despite the increase in predictive error, indicating substantial overconfidence under resolution shift. Proportional context scaling does not mitigate this calibration degradation, and in some cases further destabilizes MC-NLL. These issues mirror the deterministic transfer bottleneck: when the conditional representation does not transfer reliably across discretizations, the associated uncertainty calibration also fails to adapt appropriately to new resolution regimes.

\subsection{Local Conditioning Decomposition}
\label{sec:local_conditioning}

% =========================================================
% COMBINED TABLE 1: DETERMINISTIC LOCAL CONDITIONING DECOMPOSITION 
% (Combined from Tables 28 & 44)
% =========================================================
\begin{figure}[t]
\centering
\includegraphics[width=0.9\linewidth]{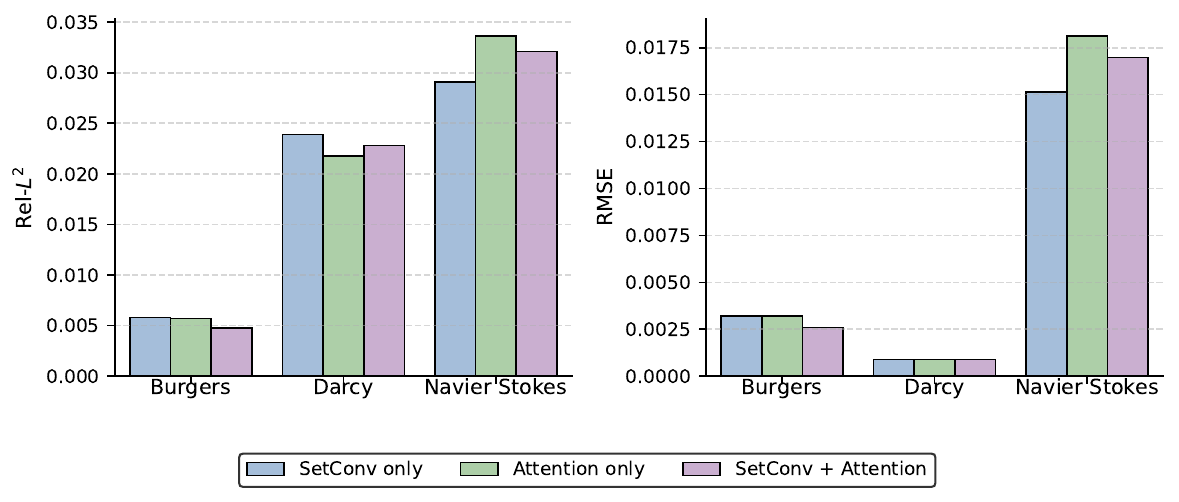}
\caption{\textbf{Deterministic local conditioning decomposition across benchmarks.} The plot shows the isolated and combined effects of the SetConv and query-aligned attention pathways within the local conditioning trunk. Mean values over 4 seeds. Lower is better.}
\label{fig:local_cond_ablation}
\vspace{-1em}
\end{figure}

%\begin{table}[htbp]
%\centering
%\caption{Deterministic local conditioning decomposition across benchmarks. Models evaluate the isolated and combined effects of the SetConv and query-aligned attention pathways within the local conditioning trunk. Mean values over 4 seeds. Lower is better.}
%\label{tab:local_cond_det_combined}
%\resizebox{0.55\linewidth}{!}{%
%\begin{tabular}{llcc}
%\toprule
%\textbf{PDE} & \textbf{Variant} & \textbf{Rel-$L^2$} & \textbf{RMSE} \\
%\midrule
%\multirow{3}{*}{Burgers} 
%& SetConv only & 0.0058 & 0.0032 \\
%& Attention only & 0.0057 & 0.0032 \\
%& SetConv + Attention & 0.0048 & 0.0026 \\
%\midrule
%\multirow{3}{*}{Darcy} 
%& SetConv only & 0.0239 & 0.0009 \\
%& Attention only & 0.0218 & 0.0009 \\
%& SetConv + Attention & 0.0228 & 0.0009 \\
%\midrule
%\multirow{3}{*}{NS} 
%& SetConv only & 0.02906 & 0.01515 \\
%& Attention only & 0.03364 & 0.01813 \\
%& SetConv + Attention & 0.03209 & 0.01698 \\
%\bottomrule
%\end{tabular}
%}
%\end{table}

% =========================================================
% COMBINED TABLE 2: PROBABILISTIC LOCAL CONDITIONING DECOMPOSITION 
% (Combined from Tables 29 & 45)
% =========================================================
\begin{table}[htbp]
\centering
\caption{Probabilistic local conditioning decomposition across benchmarks under prior-sample evaluation. The table reports predictive and calibration metrics for heteroscedastic and homoscedastic heads across all three local conditioning architectures. Mean values over 4 seeds. Lower is better for rel-$L^2$, RMSE, MC-NLL, and Width; coverage is best when close to the nominal 0.95 level.}
\label{tab:local_cond_prob_combined}
\resizebox{0.95\linewidth}{!}{%
\begin{tabular}{lllccccc}
\toprule
\textbf{PDE} & \textbf{Variance} & \textbf{Variant} & \textbf{Rel-$L^2$} & \textbf{RMSE} & \textbf{MC-NLL} & \textbf{Coverage} & \textbf{Width} \\
\midrule
\multirow{6}{*}{Burgers} 
& \multirow{3}{*}{HE} 
& SetConv only & 0.0093 & 0.0051 & -993.49 & 0.9712 & 0.0247 \\
& & Attention only & 0.0082 & 0.0046 & -1007.83 & 0.9760 & 0.0244 \\
& & SetConv + Attention & 0.0094 & 0.0052 & -994.16 & 0.9726 & 0.0248 \\
\cmidrule(lr){2-8}
& \multirow{3}{*}{HO} 
& SetConv only & 0.0114 & 0.0062 & -850.87 & 0.9862 & 0.0461 \\
& & Attention only & 0.0116 & 0.0063 & -847.02 & 0.9853 & 0.0460 \\
& & SetConv + Attention & 0.0112 & 0.0061 & -853.08 & 0.9870 & 0.0457 \\
\midrule
\multirow{6}{*}{Darcy} 
& \multirow{3}{*}{HE} 
& SetConv only & 0.0756 & 0.0030 & -16576.12 & 0.9941 & 0.0252 \\
& & Attention only & 0.0678 & 0.0027 & -16648.74 & 0.9970 & 0.0239 \\
& & SetConv + Attention & 0.0686 & 0.0027 & -16634.09 & 0.9967 & 0.0238 \\
\cmidrule(lr){2-8}
& \multirow{3}{*}{HO} 
& SetConv only & 0.0662 & 0.0026 & -16666.60 & 0.9975 & 0.0238 \\
& & Attention only & 0.0655 & 0.0026 & -16672.84 & 0.9979 & 0.0238 \\
& & SetConv + Attention & 0.0659 & 0.0026 & -16669.56 & 0.9978 & 0.0238 \\
\midrule
\multirow{6}{*}{NS} 
& \multirow{3}{*}{HE} 
& SetConv only & 0.0570 & 0.0217 & -12195.4 & 0.8982 & 0.0652 \\
& & Attention only & 0.0560 & 0.0212 & -12439.4 & 0.8971 & 0.0622 \\
& & SetConv + Attention & 0.0471 & 0.0183 & -12628.7 & 0.9017 & 0.0568 \\
\cmidrule(lr){2-8}
& \multirow{3}{*}{HO} 
& SetConv only & 0.1456 & 0.0363 & -7113.8 & 0.9159 & 0.1579 \\
& & Attention only & 0.1164 & 0.0315 & -6566.8 & 0.9195 & 0.1888 \\
& & SetConv + Attention & 0.2307 & 0.0461 & -6301.0 & 0.9245 & 0.2116 \\
\bottomrule
\end{tabular}
}
\end{table}

To isolate the contribution of each local conditioning mechanism within the geometry-aware architectures, we decompose the local trunk into three distinct configurations: a purely convolutional pathway (\textit{SetConv only}), a purely query-aligned attention pathway (\textit{Attention only}), and their combination. The objective is to determine whether the empirical gains of the final models arise primarily from continuous kernelized smoothing, dynamic query--context geometry preservation, or their fusion.

\paragraph{Deterministic Decomposition and the Role of Periodicity.} 
Figure \ref{fig:local_cond_ablation} indicates that the most effective local conditioning mechanism depends on the underlying PDE structure. On the Navier--Stokes benchmark, \textit{SetConv only} is the strongest variant in this deterministic trunk decomposition, achieving lower errors than the attention-based alternatives. In this deterministic setting, the smoother translation-friendly representation provided by SetConv is already well aligned with the downstream spectral decoder, so adding attention does not improve pure point prediction here. 

On the Burgers equation, the full \textit{SetConv + Attention} trunk performs best, although all three variants remain highly accurate. The gain is modest, but it indicates that combining the two pathways can still be beneficial even in a smooth periodic regime. By contrast, the non-periodic Darcy flow exposes a stronger role for the attention pathway. On this boundary-sensitive elliptic problem, \textit{Attention only} is the strongest deterministic configuration, slightly outperforming the fused model and clearly improving over \textit{SetConv only}. Overall, the deterministic results suggest that attention is particularly valuable when local spatial topology is critical, whereas SetConv alone can already be highly competitive in smoother periodic settings.

\paragraph{Probabilistic Decomposition and Uncertainty-Aware Conditioning.} 
Table \ref{tab:local_cond_prob_combined} shows that the role of the local conditioning pathways becomes more nuanced once uncertainty estimation is introduced. On Darcy, \textit{Attention only} remains marginally strongest, although the margin is much clearer in the HE case than in the HO. In particular, under HO variance the advantage over \textit{SetConv + Attention} is small, so the results are better interpreted as indicating that attention is the main driver of improvement on Darcy, with the fused variant remaining competitive.

A different pattern emerges on Navier--Stokes. Under HE variance, the full \textit{SetConv + Attention} trunk is clearly the best configuration, achieving the lowest rel-$L^2$ error, the best MC-NLL, and the narrowest predictive intervals. Neither pathway alone matches the combined model in this setting. This is consistent with the view that, when probabilistic prediction is required in a more demanding regime, the smoother SetConv signal and the query-aligned attention pathway can become complementary. However, this is not uniform across all settings, under HO variance on Navier--Stokes, the fused model is not the strongest configuration, and \textit{Attention only} performs better in pointwise error. The advantage of the combined trunk should also be interpreted as regime-dependent rather than universal.

We can see that the value of local geometric conditioning depends both on the benchmark and on the inference objective. \textit{SetConv only} is not uniformly sufficient, as it underperforms on Darcy and does not provide the strongest probabilistic results in the most demanding settings. \textit{Attention only} is often very strong, particularly on Darcy, but it is not consistently best across all deterministic and probabilistic regimes. The combined \textit{SetConv + Attention} design therefore remains a reasonably robust choice. Even when it is not the best performer in every individual case, it remains consistently competitive and is among the strongest configurations in several important settings, including deterministic Burgers and probabilistic Navier--Stokes with HE variance, offering a reliable overall trade-off across benchmarks and inference objectives.

\subsection{Computational Cost and Model Complexity}
\label{sec:cost_complexity}

% =========================================================
% COMBINED TABLE: COMPUTATIONAL COST AND COMPLEXITY
% (Combined from Tables 30, 31, and 41)
% =========================================================

\begin{table}[htbp]
\centering
\setlength{\tabcolsep}{2pt}
\caption{Computational cost and complexity profiling across benchmarks. Mean values across 4 seeds, measured using evaluation batch size $B=16$ for Burgers and $B=8$ for Darcy/Navier--Stokes. Probabilistic inference uses 8 Monte Carlo samples. Training costs increase with the geometry-aware and probabilistic extensions, while deployment-time latency remains modest.}
\label{tab:cost_complexity_combined}
\resizebox{\linewidth}{!}{%
\begin{tabular}{llccccc}
\toprule
\textbf{PDE} & \textbf{Model} & \textbf{Parameters} & \textbf{Infer time/batch (s) $\downarrow$} & \textbf{Peak infer mem (MB) $\downarrow$} & \textbf{Epoch time (s) $\downarrow$} & \textbf{Peak train mem (MB) $\downarrow$} \\
\midrule
\multirow{6}{*}{Burgers} 
& DNOP & 418,625 & 0.00531 & 43.31 & 0.494 & 92.62 \\
& DANOP & 456,577 & 0.00568 & 48.90 & 0.718 & 110.37 \\
& PNOP-HE & 446,594 & 0.00506 & 43.45 & 0.694 & 102.21 \\
& PNOP-HO & 446,530 & 0.00532 & 43.44 & 0.699 & 102.19 \\
& PANOP-HE & 575,298 & 0.00590 & 46.36 & 1.030 & 153.28 \\
& PANOP-HO & 575,234 & 0.00597 & 46.35 & 1.040 & 153.22 \\
\midrule
\multirow{6}{*}{Darcy} 
& DNOP & 4,220,097 & 0.00338 & 276.30 & 1.059 & 417.89 \\
& DANOP & 4,258,497 & 0.00434 & 341.74 & 1.514 & 749.78 \\
& PNOP-HE & 4,248,066 & 0.00387 & 276.66 & 2.127 & 4004.46 \\
& PNOP-HO & 4,248,002 & 0.00386 & 276.54 & 2.130 & 4004.33 \\
& PANOP-HE & 4,377,666 & 0.00468 & 277.16 & 4.331 & 4125.52 \\
& PANOP-HO & 4,377,602 & 0.00469 & 277.03 & 4.327 & 4125.40 \\
\midrule
\multirow{4}{*}{NS} 
& PNOP-HE & 4,247,746 & 0.00390 & 281.29 & 2.135 & 4004.33 \\
& PNOP-HO & 4,247,682 & 0.00388 & 281.16 & 2.152 & 4004.20 \\
& PANOP-HE & 4,376,706 & 0.00501 & 281.78 & 4.365 & 4124.64 \\
& PANOP-HO & 4,376,642 & 0.00485 & 281.65 & 4.355 & 4124.51 \\
\bottomrule
\end{tabular}
}
\end{table}

We note the main asymptotic distinction between the conditioning and decoding stages. For $n_C$ context points and $n_Q$ query locations, both SetConv and query-aligned attention require evaluating context--query interactions, yielding $O(n_C n_Q)$ computational cost and $O(n_C n_Q)$ memory when the corresponding interaction weights are materialized. The FNO decoder instead scales as $O(L n_Q \log n_Q)$ in the query grid size for $L$ spectral layers, up to channel, mode, and implementation constants. Thus, the conditioning stage is inexpensive in the sparse regimes considered here, where $n_C \ll n_Q$, but can become a bottleneck as the context set grows.

To assess whether the empirical gains of the geometry-aware and probabilistic extensions come at a reasonable computational cost, we profile the models' resource requirements across all three benchmarks. The objective is to quantify the overhead introduced by the query-aligned attention pathway and the latent conditioning mechanisms relative to the SetConv-only deterministic baseline (\textsc{DNOP}). Table \ref{tab:cost_complexity_combined} summarizes the parameter counts, inference latency per batch, peak inference memory, mean epoch training time, and peak training memory for the final model configurations. For probabilistic models, inference is measured under the prior-sampled setting with a fixed Monte Carlo budget, reflecting the actual deployment-time evaluation.

\paragraph{The Training vs. Inference Cost Asymmetry.} 
The main pattern emerging across benchmarks is that computational cost increases in a predictable order with architectural complexity, but this overhead is asymmetric; it is observed primarily during training rather than at inference. On Burgers, the inference latency of all models remains tightly clustered between $5.0$ and $6.0$ milliseconds per batch. Similarly, on Darcy, while inference time grows from the baseline \textsc{DNOP} ($\sim 3.38$ ms) to the full probabilistic \textsc{PANOP} ($\sim 4.69$ ms), it remains strictly on the order of milliseconds. Thus, deployment-time latency stays low across all profiled models, even as training cost increases.

By contrast, the training cost scales much more sharply. On Darcy, \textsc{DANOP} increases the mean epoch time by roughly $43\%$ over \textsc{DNOP}, and the latent probabilistic models increase it by factors of roughly two to four. This trend is even more pronounced in peak training memory: while the deterministic models remained below 1 GB on Darcy, the probabilistic models required approximately 4 GB. This jump is much larger than what parameter count alone would suggest. The main practical cost of the probabilistic operator-process models may therefore lie in the optimization graph and training-time activations required by latent inference, rather than in deployment-time execution.

\paragraph{Computational Graph vs. Parameter Count.} 
The Navier--Stokes profiling data isolates the cost differences between the probabilistic variants and helps clarify the source of the overhead. \textsc{PANOP} is only marginally larger in parameter count than \textsc{PNOP} (an increase of approximately $3\%$) and exhibits nearly identical peak memory usage during both inference and training. However, its runtime overhead is pronounced: mean epoch time roughly doubles and inference time per batch increases by approximately $25$--$28\%$. The computational premium of \textsc{PANOP} is not well explained by parameter count alone, but appears to be driven more by the computational graph induced by the architecture, specifically, the evaluation and fusion of both SetConv and query-aligned attention pathways before the operator decoder. The cost gap is therefore driven primarily by how the model computes, rather than by a large increase in raw parameter capacity.

\paragraph{Inference Memory Profiling.}
Peak inference memory is implementation-dependent, reflecting batching, activation materialization, CUDA allocator behavior, and evaluation scope. We therefore treat it as a practical profiling statistic rather than a strict architectural ordering. The conclusion is that probabilistic variants mainly increase training-time cost, while inference latency remains modest.

These analyses support a favorable cost--performance trade-off for the proposed architectures, with an important qualification: the additional cost is concentrated mainly in training rather than in inference. While \textsc{PANOP} is substantially more expensive to optimize, its parameter count remains close to the simpler baselines, its inference latency stays within a small millisecond-scale range, and its deployment-time memory footprint remains manageable. Given the gains in geometric robustness and the improved probabilistic behavior observed in the main evaluations and related ablations, this additional optimization cost appears reasonable for applications requiring reliable operator inference under sparse partial observations.

\end{document}